\documentclass[a4paper, 12pt]{article}

\usepackage{lineno,hyperref}
\DeclareMathAlphabet{\pazocal}{OMS}{zplm}{m}{n}
\newcommand{\Lb}{\pazocal{X}}
\usepackage{dsfont}
\usepackage{amsmath}
\usepackage{tcolorbox}
\usepackage{subfig}
\usepackage{float}
\usepackage{longtable}
\usepackage{booktabs}
\usepackage{lscape}
\usepackage[flushleft]{threeparttable}
\usepackage{multirow}
\usepackage{afterpage}
\usepackage{hyperref}
\usepackage{natbib}
\hypersetup{
    colorlinks=true,
    filecolor=magenta,      
    citecolor=blue,
}

\modulolinenumbers[5]

\def\argmax{\mathop{\rm arg\,max}\limits}

\newtheorem{theorem}{Theorem}[section]
\newtheorem{definition}[theorem]{Definition}

\newtheorem{proposition}[theorem]{Proposition}

\newtheorem{axiom}{Axiom}

\begin{document}


\title{Clustering with the Average Silhouette Width}


\author{Fatima Batool (1) and Christian Hennig (2)\\
(1) Department of Statistical Science\\
       University College London\\
       Gower Street, London WC1E 6BT, United Kingdom\\
(2) Dipartimento di Scienze Statistiche ``Paolo Fortunati''\\
       Universita di Bologna\\
       Bologna, Via delle belle Arti, 41, 40126, Italy}

       
\maketitle

\begin{abstract}
The Average Silhouette Width (ASW; \citet{rousseeuw1987silhouettes}) is a popular cluster validation index to estimate the number of clusters. Here we address the question whether it also is suitable as a general objective function to be optimized for finding a clustering. We will propose two algorithms (the standard version OSil and a fast version FOSil) and compare them with existing clustering methods in an extensive simulation study covering the cases of a known and unknown number of clusters. Real data sets are also analysed, partly exploring the use of the new methods with non-Euclidean distances. We will also show that the ASW satisfies some axioms that have been proposed for cluster quality functions (\citet{ben2009measures}). The new methods prove useful and sensible in many cases, but some weaknesses are also highlighted. These also concern the use of the ASW for estimating the number of clusters together with other methods, which is of general interest due to the popularity of the ASW for this task.   

{\bf Keywords:} axiomatic clustering, distance-based clustering,  partitioning around medoids, number of clusters\\ 
MSC2010 classification: 62H30
\end{abstract}

\section{Introduction}
Cluster analysis is a central task in modern data analysis. It is applied in diverse areas such as public health, machine learning, psychology, archaeology, genetics, computer vision, and text analytics to just name a few. There is a wide variety of clustering methods (\citet{hennig2015handbook}), and it has been argued that there is no universally best approach, and that the cluster analysis approach needs to be chosen taking into account what kinds of clusters are required, which depends on domain knowledge and the aim of clustering (\citet{von2012clustering,hennig2015true}). 

Many cluster analysis methods such as $k$-means (\citet{lloyd1982least}) and Partitioning Around Medoids (PAM; \citet{kaufman1987clustering}) are defined by optimizing an objective function over all partitions of the data set into a fixed number of $k$ clusters. The objective function formalizes the quality of the clustering. The PAM objective function, for example, sums up distances of all observations to the center of the cluster to which they were assigned, and a good clustering is one for which this is small. Many of these objective functions are not suitable for finding an optimal number of clusters $k$, because they will automatically improve if $k$ is increased due to more degrees of freedom in optimization. 

Therefore, so-called cluster validation indexes that can be meaningfully optimized over $k$ are often used together with such partitioning clustering methods in order to find an optimal $k$. Many such indexes have been proposed in the literature (see  \citet{hennig2015clustering26,AGMPP12}). Some of these are for fixed $k$ equivalent to objective functions of partitioning methods such as $k$-means, e.g., the Calinski and Harabasz index (\citet{calinski1974dendrite}). Some others are not in this way connected to a specific partitioning method. One popular such index is the Average Silhouette Width (ASW) (\citet{rousseeuw1987silhouettes}). The ASW achieved overall very good results in the extensive simulation study of \citet{AGMPP12}. In  \citet{kaufman1990finding} it is suggested for finding the number of clusters with PAM, but in fact its definition is not directly connected to any specific partitioning method, and it can be seen as a general distance-based approach to assess the quality of a clustering. 

The ASW is an intuitive and simple measurement of cluster quality that does not rely on statistical model assumptions. Given that it is widely used and trusted for comparing the qualities of clusterings produced by various clustering methods over different numbers of clusters, it seems natural to investigate optimal ASW quality clustering not only over $k$ but also for fixed $k$ in order to integrate the problem for fixed $k$ and the problem of finding the best $k$. This idea is explored here. We treat the idea with an open mind and do not attempt to suggest that optimum ASW clustering is an optimal clustering method in any other sense than optimizing the ASW (which can be seen as desirable on its own terms); rather what we do is to show both potential and problems with this approach. The problems are also relevant to the use of the ASW for just choosing $k$, for which it is of widespread use despite a far from comprehensive evaluation and theoretical basis. To our knowledge, up to now using the ASW also for choosing a clustering with fixed $k$ has only been explored by \citet{van2003new}, where a modification of the PAM algorithm called PAMSil was proposed that looks for a local medoid-based optimum of the ASW. \citet{rousseeuw1987silhouettes}, where the ASW was originally introduced, mentions a possibility of its optimisation for finding a clustering in a side remark. Exploring the new clustering method based on the ASW, we also to some extent explore strengths and weaknesses of the popular use of the ASW as a method to estimate the number of clusters.

In Section \ref{smethodology} we introduce optimum ASW clustering and propose two algorithms for it. 
In Section \ref{saxioms} we show that the ASW fulfils some axioms that have been proposed for clustering quality measures in \citet{ben2009measures}. 
In Section \ref{ssimulation} we run an extensive simulation study to explore the performance of optimum ASW clustering compared to other well established clustering methods. Section \ref{sappl} applies optimum ASW clustering to a number of real data sets with and without given ``true'' clustering, also illustrating the use of optimum ASW for non-Euclidean dissimilarities. A conclusion is given in Section \ref{sconclusion}.

\section{Methodology} \label{smethodology}
\subsection{Notation and basic definitions}
Let $X =  \{x_1, \dots, x_n\} $ be a data set of $n$ objects from a space $\Lb$, $d$ be a dissimilarity or distance over $\Lb$. The triangle inequality is not really necessary here, although the intuition behind the concepts cluster separation and homogeneity may look dubious if for example for points $x_1, x_2, x_3$ it is possible that $d(x_1,x_2)$ and $d(x_2,x_3)$ are very small, but $d(x_1,x_3)$ is very large. We deal with clusterings that are partitions, i.e., non-overlapping and exhaustive. A partition can equivalently be expressed by labels $l(1), \dots, l(n) \in \mathds{N}_k=\{1,\ldots,k\}$ where $l(i) = r\Leftrightarrow x_i\in C_r$,  $i \in \mathds{N}_n,$ and cluster sizes are denoted by $n_r = \sum_{i=1}^{n} 1 (l(i) = r)$, $r \in \mathds{N}_k$. 
\begin{definition}
The silhouette width for an observation $x_i\in X$ is 
\begin{equation}\label{equations}
s_i(\mathcal{C}, d) = \frac{b(i)  - a(i)} { \textrm{max} \{ a(i), b(i)\}},
\end{equation}
where 
\[
a(i) =  \frac{1}{n_{{l(i)}}-1} \sum_{\substack{  l(i) = l(j)\\
                 i \neq j }} d(x_i, x_j)
\quad \text{and} \quad
b(i) =  \min_{r \neq l(i)} \frac{1}{n_r} \sum_{l(j) = r} d(x_i, x_j)
\]
in case that $n_r>1$ for $l(i)=r$. Otherwise $s_i(\mathcal{C}, d)=0$.

The \textbf{Average Silhouette Width} (ASW) of a clustering $\mathcal{C}$ is
\begin{equation*}
\bar{S}(\mathcal{C}, d) = \frac{1}{n} \sum_{i=1}^n s_i(\mathcal{C}, d).
\end{equation*}
\end{definition}
$a(i)$ is the average distance of $x_i$ to points in the cluster to which it
was assigned, and $b(i)$ is the average distance of $x_i$ to the points in the 
nearest cluster
to which it was not assigned. A large value of $s_i(\mathcal{C}, d)$ means that
$b(i)$ is much larger than $a(i)$, and that consequently $x_i$ is much closer to the observations in its own cluster than to the neighboring one. Given that clusters are meant to be homogeneous and well separated, larger values of $s_i$ and $\bar{S}$ indicate better clustering quality, and an optimal clustering (for example with optimal $k$ if various values of $k$ are compared) in the sense of the ASW is one that maximizes $\bar{S}$. A more in-depth heuristic motivation of silhouette widths along with some examples is given by \citet{rousseeuw1987silhouettes}, which focuses on the graphical display of individual silhouette widths but also introduces the ASW for assessment of the whole clustering. 

Here are some simple properties of the ASW.
$-1\le s_i\le 1$ always, and the same holds obviously for the ASW. In fact an ASW of 0 can be seen as a rather bad value, because it means that on average observations are not closer to the observations in their own cluster than to the observations in the closest other cluster. However, a random clustering with more than 2 clusters cannot normally be expected to achieve $\bar{S}=0$, because $b(i)$ is computed by minimizing over clusters to which an observation does not belong, and on average this minimum can be expected to be smaller than $a(i)$ if the observations in the same cluster are as randomly chosen as those in the other clusters. 

If there are well separated and compact subsets in the data, taking these as the clusters will make the vast majority of $s_i$ and consequently $\bar{S}$ substantially larger than zero, and this will be better than having $k$ close to its maximum value $n$ (in which case many one-point clusters will lead to $s_i=0$ for many $i$). Putting two or more such subsets together in a cluster will have a detrimental impact on the corresponding $a(i)$-values, damaging $\bar{S}$ in turn, so that the optimum value of $\bar{S}$ will not normally occur at a $k$ lower than the number of well separated subsets either. There is a possible exception to this though. Putting two neighboring clusters together can make their separation from the rest, i.e., the corresponding $b(i)$-values, much higher, so that occasionally data subsets that have some separation from each other put together produce a better ASW-value, if their separation from the rest is much stronger. This can be seen as a general pitfall of the ASW, particularly when it comes to estimating $k$. See also \citet{hennig15paramboot}, where maximizing the ASW over $k$ (for given $k$, clusterings were produced by PAM) produces an apparently too low value of $k=2$. 

The ASW cannot be computed for $k=1$, so it cannot directly be used to decide whether the data set as a whole is homogeneous and whether there should be any clustering at all. \citet{hennig15paramboot} suggest to compare ASW values on the data to ASW values from clustering homogeneous ``null model'' data without clustering to see whether the data have a significant clustering structure. 
\begin{definition}
A clustering $\mathcal{C}^*$ is an {\bf optimum ASW clustering} for dissimilarity $d$ if
\begin{equation}\label{saswdef}
  \bar{S}(\mathcal{C}^*,d)=\max_{\mathcal{C}} \bar{S}(\mathcal{C},d), 
\end{equation}
where $\mathcal{C}\in\mathcal{P}(X)$ with $|\mathcal{C}|\ge 2$. Define $\mathcal{C}^*_k$ as optimum ASW clustering out of the clusterings with $|\mathcal{C}|=k$.
\end{definition}
As with other clustering principles that are defined as optimizing an objective function, finding a global optimum will be computationally infeasible for all but the smallest data sets. We therefore propose two algorithms to find local optima that are hopefully close to the global one.

\subsection{The OSil algorithm} \label{sosila}
The following algorithm improves an initialisation by changing the cluster membership of the point that improves the ASW most at any given stage until no further improvement can be found. As the ASW will always be improved and there are only finitely many clusterings, the algorithm will converge in a finite number of steps. We call this the OSil (optimum silhouette) algorithm.

\begin{minipage}{14.5cm}
\noindent \rule[2pt]{\linewidth}{2pt}\vspace{-.5mm}\\
\textbf{OSil algorithm}\\
\noindent \rule[2ex]{14.5cm}{0.90pt} \\
{\fontsize{10}{10} \selectfont
\begin{itemize}
\item Set $k_{min}$ as the minimum number of clusters and $k_{max}$ as the 
maximum number of clusters.  
\item Input: dissimilarity $d(x_i, x_h)$, $\forall i \neq  h \in \mathds{N}_n$. 
\item For every $k\in\{k_{min},\ldots,k_{max}\}$:
\begin{enumerate}  
\item Start with initialisation clustering ${\mathcal C}_k^0$ defined by $l^0(\Lb, k) = (l^0(1), \dots, l^0(n))$, for $i\in\mathds{N}_n:\ l^0(i)\in\mathds{N}_k.$
Let $q=0$.
\item Compute $\bar{S}(\mathcal{C}_k^q, d)$.
\item  For all pairs $(i, r)$ such that $l^0(i) \neq r$, $i \in \mathds{N}_n,\ r \in \mathds{N}_k$, assign $l^{(i, r)}(i)=r, \ l^{(i, r)}(j)=l^0(j)$ for all $j\neq i$, and denote the so obtained clustering as ${\mathcal C}_k^{(i,r)}$. 
\item  Compute $f^{(i,r)} = \bar{S}(\mathcal{C}_k^{(i, r)}, d)$. 
\item  $(h, s) = \underset{(i, r) }{\arg\max} ~~ f^{(i, r)} $ (we recommend to constrain $(i,r)$ so that $\mathcal{C}_k^{(i, r)}$ still has $k$ nonempty clusters, but not using this constraint would be an alternative if for given $k$ a clustering with $k$ nonempty clusters is not necessarily required), 
\item If $f^{(h, s)}\le \bar{S}(\mathcal{C}_k^q, d)$: $q=q+1$,
$\mathcal{C}_k^{(q)}=\mathcal{C}_k^{(h, s)}$, and go to Step 2. Otherwise stop
and give out $\mathcal{C}_k=\mathcal{C}^{(q)}_k$ as final solution for number of clusters $k$.
\end{enumerate}
\end{itemize}
Give out 
$\mathcal{C}_{k^*}=argmax_{k\in\{k_{min},\ldots,k_{max}\}} \bar{S}(\mathcal{C}_k, d)$ as final clustering.
}\\
\noindent\rule[2ex]{\linewidth}{0.90pt}
\end{minipage}

It is advisable to run the algorithm more than once from different initialisations, and then to use the solution that achieves the best ASW-value. We ran some simulations comparing different possible ways of initialisation, particularly initialisation by already existing clustering methods, see \citet{batool2019initial}. Good solutions over a variety of setups can be achieved initializing six times with $k$-means, PAM, average linkage, single linkage, Ward's method and model-based clustering (using the standard settings of the R-package \verb|mclust| (\citet{fraleymclust}), which we recommend here.

\subsection{FOSil - An approximation algorithm for bigger data sets}
The OSil algorithm is computationally expensive, because it considers all possible combinations of cluster and observation swaps for each iteration. For large data sets it can be very slow. A simple way to construct a faster algorithm for larger data sets is to run the OSil algorithm on a subset of observations, and then to assign all remaining observations to the clusters in such a way that the ASW is maximized for every observation separately. We call this the FOSil (Fast OSil) algorithm.

\begin{minipage}{14.2cm}
\noindent \rule[2pt]{\linewidth}{2pt}\vspace{-.5mm}\\
\textbf{FOSil algorithm}\\
\noindent \rule[2ex]{14.2cm}{0.80pt} \\
{\fontsize{9}{9} \selectfont
\begin{itemize}
\item Set $k_{min}$ as the minimum number of clusters and $k_{max}$ as the 
maximum number of clusters. 
\item Set a sample size $n_s$ and a number $M$ of samples to be drawn.
\item Input: dissimilarity $d(x_i, x_h)$, $\forall i \neq  h \in \mathds{N}_n$. 
\item For every $k\in\{k_{min},\ldots,k_{max}\}$:
\begin{enumerate}  
\item For $m=1,\ldots,M$:
\begin{enumerate}
\item Choose a random sample $S_m$ of size $n_s$ from $X$. Let $I_{S_m}\subset\{1,\ldots,n\}$ be the set of indexes of the observations in $S_m$. Let $d_m$ be $d$ constrained to the elements of $S_m$. 
\item Run the OSil algorithm on $S_m, d_m$ with number of clusters $k$, potentially starting from several initialisations (see the discussion at the end of Section \ref{sosila}). Call the resulting clustering ${\mathcal C}_{S_m,k}$. For a label vector $(l_{S_m,k}^*(1),\ldots,l_{S_m,k}(n))$ define $l_{S_m,k}(i)=r$ if $x_i\in C_r$ in ${\mathcal C}_{S_m,k}$ for $i\in I_{S_m}$.
\end{enumerate}
\item Choose $S=\argmax_{S_m\in\{S_1,\ldots,S_M\}}\bar{S}(S_m,d_m)$.  
\item Calculate the cluster memberships for the points in $S'=X\setminus S$ by maximizing the ASW. For all $x_i\in S'$ and $C_r\in{\mathcal C}_{S,k},\ i\in \{2,\ldots,n\}\setminus I_S,\ r\in\{1,\ldots,k\}$:
  \begin{enumerate}
  \item Consider the clustering 
$\mathcal{C}_{S,k}^{(i,r)}$ of $S\cup\{x_i\}$ defined by putting $x_i\in C_r$ and otherwise leaving ${\mathcal C}_{S,k}$ unchanged. Let $d^*$ denote $d$ constrained to the elements of $S\cup\{x_i\}$.
  \item Compute  $f^{(i,r)} = \bar{S}(\mathcal{C}_{S,k}^{(i, r)}, d^*)$.
  \item $(h, s) = \underset{(i, r) }{\arg\max} ~~ f^{(i, r)}$; $l_{S,k}(h)=s$.   
  \end{enumerate}
\item Give out $C_k$ as defined by $l_{S,k}$ as the final solution for number of clusters $k$.
\end{enumerate}
\end{itemize}
Give out 
$\mathcal{C}_{k^*}=argmax_{k\in\{k_{min},\ldots,k_{max}\}} \bar{S}(\mathcal{C}_k, d)$ as final clustering.
}\\
\noindent\rule[2ex]{\linewidth}{0.90pt}
\end{minipage}

\begin{figure}[H]
\centering
\includegraphics[width=0.48\textwidth]{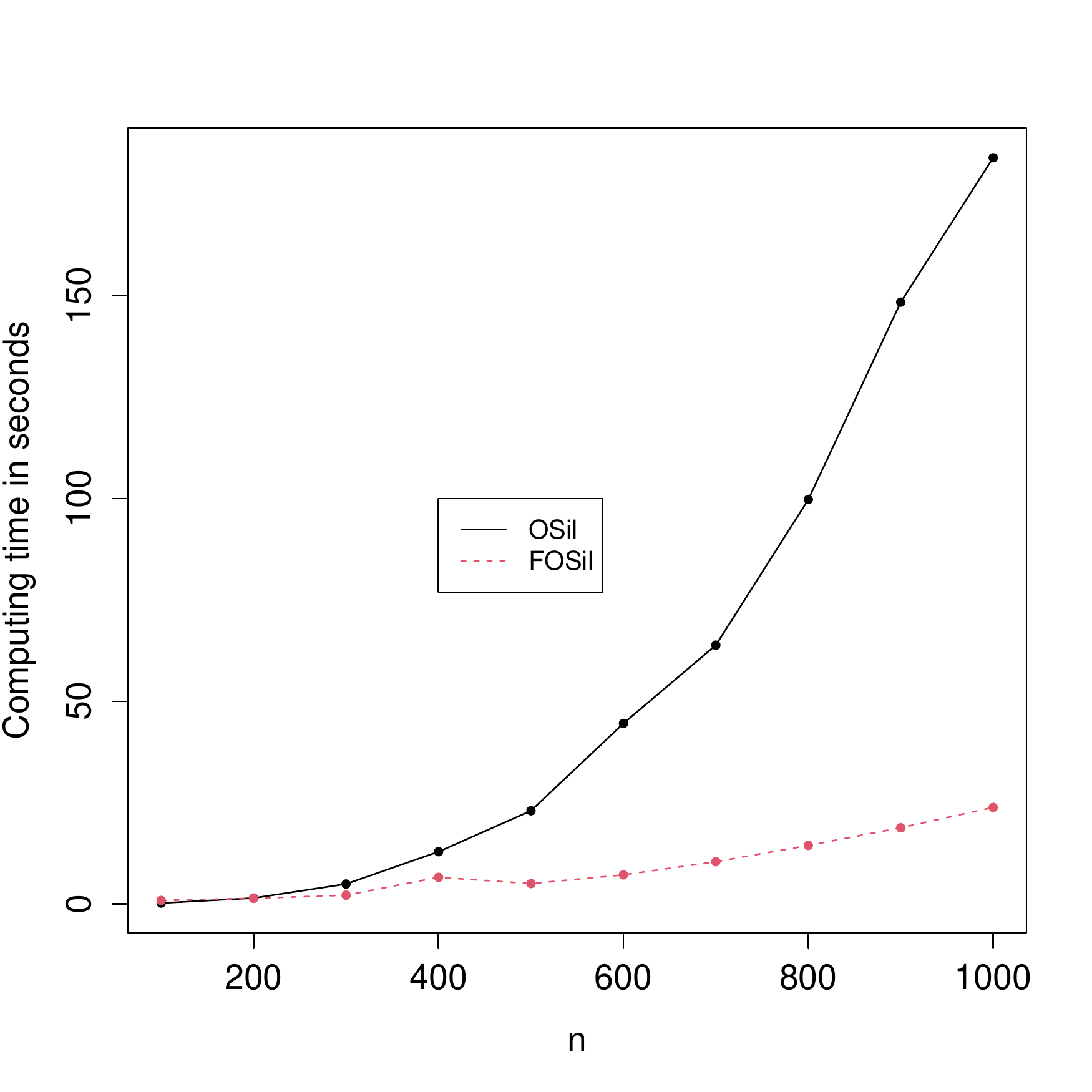}
\caption{Computing times of OSil and FOSil on two-dimensional data sets with equally sized Gaussian clusters centered at (0,0), (0,1), (1,0), and (1,1), with standard deviations 0.1 in both independent dimensions for $n$ between 100 and 1000.}
\label{ftimes}
\end{figure}

We used $M=25$ (larger $M$ does not seem to improve matters much) and $n_s=0.2n$, although for larger data sets $n_s$ can be chosen smaller; the absolute size of $n_s$ will often matter more than its ratio to $n$. Alternatively, one could choose 20 observations times the maximum number of clusters of interest.
As in Section \ref{sosila}, we initialize the clustering of the $S_m$ by  $k$-means, PAM, average linkage, single linkage, Ward's method and model-based clustering.
See Figure \ref{ftimes} for a comparison of computing times of OSil and FOSil.

There are some possible variations of the FOSil algorithm. For example, when assigning the points in $S'$ to the clusters, one may use for classifying a new point the ASW including all points already assigned rather than just $S$ plus the new point. Also, points in $S'$ may be assigned to the clusterings ${\mathcal C}_{S_m,k}$ from all $M$ subsamples before comparing the results from the different subsamples. Both of these can be expected to improve the clustering but will take more computation time than FOSil. We do not explore them here.

\section{Axiomatic characterisation of the ASW} \label{saxioms}

The ASW has originally been introduced as a heuristic concept. Despite its popularity and good results in some studies, to this day there has not been much theoretical investigation of its characteristics. We apply an axiomatic approach by \citet{ben2009measures} to the ASW. The general idea is to characterize a desirable behavior of a reasonable clustering method (or clustering quality measure (CQM)) by certain theoretical axioms and then to check whether for a given method or CQM these axioms are fulfilled. This approach goes back to at least \citet{rubin1967optimal}. Some further early work was done by \citet{jardine1968construction}, \citet{fisher1971admissible}, and \citet{wright1973formalization}.

More recently, \citet{kleinberg2003impossibility} defined three apparently intuitive axioms for clustering functions (i.e., clustering methods), ``scale invariance'' (the clustering does not change if all dissimilarities are multiplied by the same constant), ``consistency'' (if dissimilarities are changed in such a way that all within-cluster dissimilarities are either made smaller or unchanged, and all between-cluster dissimilarities are either made larger or unchanged, the clustering does not change), and ``richness'' (for all possible clusterings there is a dissimilarity that will produce them). He then proved that it is impossible for any clustering function to fulfil all three of them. \citet{ben2009measures} argued that particularly the consistency axiom is not as desirable as \citet{kleinberg2003impossibility} had claimed. They proposed versions of the three axioms that do not apply to clustering functions but rather to CQMs such as the ASW, which we will use here. In this way, all three can be fulfilled. Some other relaxations of Kleinberg's axioms have been proposed by  \citet{zadeh2009uniqueness}, \citet{correa2013indication}, and \citet{carlsson2013classifying}.

\subsection{Definitions and axioms}

Let  $X = \{ x_1,\cdots,x_n\}$ be the data set as before and $d$ a dissimilarity on $X$.
For a given partition $\mathcal{C}$ of $X$, write $ x_i \sim_{\mathcal{C}} x_j$ if $x_i$ and $x_j$ are in the same cluster in $\mathcal{C}$. Clustering $\mathcal{C}$ is called non-trivial if not either $\mathcal{C}=\{X\}$ (only the whole data set is a cluster) or $\mathcal{C}=\{\{x_1\},\ldots,\{x_n\}\}$ (all $n$ singletons are the clusters). A clustering quality measure (CQM)  $\Pi$ takes the pair $(X, d)$ and a clustering $\mathcal{C}$ over $(X, d)$ and returns a  non-negative real number, where a larger number is interpreted as a higher cluster quality.

\begin{definition} \label{alphadist}
For a dissimilarity $d$ over $X$ and a positive real $\eta$, the scalar multiplication of $d$ with $\eta$ is defined for every pair $x_i, x_j \in X$, as ($\eta \cdot d$ )($x_i$, $x_j$) = $\eta d(x_i, x_j)$.
\end{definition}

\begin{definition}\label{ctrans}
A dissimilarity $d^\prime$ is called a $\mathcal{C}$-\textbf{transformation} of $d$, if $d^\prime(x_i, x_j) \leq d(x_i, x_j)$ for all  $ x_i \sim_{\mathcal{C}} x_j$ and $d^\prime(x_i, y_j) \geq d(x_i, x_j)$ for all  $ x_i \not\sim_{\mathcal{C}} x_j$ for all $i, j \in \mathds{N}_n$.
 \end{definition}
\begin{definition}
Two clusterings $\mathcal{C}$ and $\mathcal{C}'$ of $X$ are \textbf{isomorphic} ($\mathcal{C}\approx_d\mathcal{C}'$ if there exists a distance-preserving isomorphism $\phi:\ X\mapsto X$,such that for all $x, y\in X:\ x \sim_{\mathcal{C}} y\Leftrightarrow \phi(x)\sim_{\mathcal{C}'}\phi(y)$.
\end{definition}
 
\citet{ben2009measures} state their axioms as follows. 
\begin{axiom} 
\textbf{CQM scale invariance:} A CQM $\Pi$ is scale invariant if for all $\eta > 0$, and every $\mathcal{C}$ of ($X$, $d$), $\Pi(\mathcal{C}, (X, d)) = \Pi(\mathcal{C}, (X, \eta \cdot d))$. 
\end{axiom}

\begin{axiom}
\textbf{CQM consistency:}  A CQM $\Pi$ is  consistent  if for every clustering  $\mathcal{C}$ over ($X$, $d$),  $\Pi(\mathcal{C}, (X, d^\prime)) \geq \Pi(\mathcal{C}, (X, d))$ holds, provided that $d^\prime$ is a $\mathcal{C}$-transformation of $d$. 
\end{axiom}

\begin{axiom}
\textbf{CQM richness:} A CQM $\Pi$ is rich for every possible  non-trivial clustering $\mathcal{C} \in \mathcal{S}(X)$ of $X$ there exist a dissimilarity $d$ over $X$ such that $\mathcal{C} = \arg\max_\mathcal{C}\Pi(\mathcal{C}, (X, d))$.
\end{axiom}
\begin{axiom}
\textbf{Isomorphism invariance:} A CQM $\Pi$ is isomorphism-invariant if for allclusterings $\mathcal{C},\mathcal{C}'$ over $(X,d)$ where $\mathcal{C}\approx_d\mathcal{C}':\ \Pi(\mathcal{C}, (X, d)) =\Pi(\mathcal{C}', (X, d))$.
\end{axiom}

If a clustering method is defined by maximizing a CQM (such as the ASW),
CQM richness is identical to the richness axiom in 
\citet{kleinberg2003impossibility}, and CQM scale invariance implies the scale invariance axiom in \citet{kleinberg2003impossibility}. CQM consistency is weaker than \cite{kleinberg2003impossibility}'s consistency for clustering methods.

The axioms are justified as follows. Scalar multiplication should not affect the grouping structure of a data set, as all dissimilarities are modified in the same way. Regarding CQM consistency, a $\mathcal{C}$-transformation improves both the within-cluster homogeneity and the between-cluster separation, so the existing clustering should be rated as of higher quality by the CQM. \citet{kleinberg2003impossibility}'s consistency demands more, but it is hard to justify that the transformation is not allowed to lead to another even better clustering, see \citet{ben2009measures}. The rationale behind richness is that if any non-trivial clustering cannot be achieved by constructing a dissimilarity for which this clustering is optimal, non-optimality of that clustering is an artefact of the CQM (or the clustering method) rather than a defect of the clustering itself. \citet{ben2009measures} show that CQM scale invariance, CQM richness, and CQM consistency form a consistent set of axioms. They added isomorphism invariance in order to stop some functions that would be unreasonable as CQMs from fulfilling all axioms.

\subsection{Characterisation of the ASW} \label{scharacter}
We now prove CQM scale invariance, CQM richness, and  CQM consistency for the ASW.  Isomorphism invariance is trivially fulfilled because the ASW depends on 
dissimilarities only.

\begin{proposition}
The ASW is a scale invariant CQM.
\end{proposition}
{\bf Proof:} If $d$ is replaced by $\eta\dot d$, for all $i$ both $a(i)$ and
$b(i)$ are multiplied by $\eta$, and therefore $s_i$ does not change.
\begin{theorem}
The ASW is a consistent CQM.
\end{theorem}
{\bf Proof:} 
 Let $d^\prime$ be a $\mathcal{C}$-transformation  of $d$, and $a^\prime(i)$, $b^\prime(i)$, $s_i^\prime=s_i(\mathcal{C}, d^\prime)$, $\bar{S}^\prime=\bar{S}(\mathcal{C}, d^\prime)$ denote the corresponding quantities from the definition of the ASW based on $d^\prime$. By Definition \ref{ctrans}: 
$d^\prime(x_i, x_j) \leq d(x_i, x_j)$  for all  $ x_i  \sim_{\mathcal{C}} x_j$, and
 $\min_{ x_i \not\sim_{\mathcal{C}} x_j}d^\prime(x_i, y_j) \geq \min_{ x_i \not\sim_{\mathcal{C}} x_j} d(x_i, x_j)$. This implies for all $i\in\mathds{N}_n:$
\begin{equation} \label{eqaprime} 
a^\prime(i) \leq a(i),\ b^\prime (i) \geq b(i).
\end{equation}
Show for all $i$:
\begin{equation}\label{maineqcont}
\frac{b^\prime(i) - a^\prime(i)}{\textrm{max}\{a^\prime(i), b^\prime(i)\}} - \frac{b(i) - a(i)}{\textrm{max}\{a(i), b(i)\}} \geq 0,
\end{equation}
which is equivalent $s_i^\prime(\mathcal{C}, d) \geq s_i(\mathcal{C}, d),$ and 
implies CQM consistency, as $\bar{S}^\prime$ and $\bar{S}$ average these.

There are four possible cases:
\begin{align}
&\text{Case I:} ~\textrm{max}\{a(i), b(i)\} = a(i), 
&\textrm{max}\{a^\prime(i), b^\prime(i)\} = a^\prime(i) \label{equ1} \\
&\text{Case II:} ~ \textrm{max}\{a(i), b(i)\} = a(i), 
&\textrm{max}\{a^\prime(i), b^\prime(i)\} = b^\prime(i) \label{equ2}\\
&\text{Case III:} ~ \textrm{max}\{a(i), b(i)\} = b(i), 
&\textrm{max}\{a^\prime(i), b^\prime(i)\} = a^\prime(i) \label{equ3}\\
&\text{Case IV:} ~ \textrm{max}\{a(i), b(i)\} = b(i), 
& \textrm{max}\{a^\prime(i), b^\prime(i)\} = b^\prime(i). \label{equ4}
\end{align}
Check whether (\ref{maineqcont}) holds for each of these:\\
\textbf{Case I:} (\ref{maineqcont}) amounts to
\begin{align}\label{concondone}
&\frac{b^\prime(i) - a^\prime(i)}{a^\prime(i)} - \frac{b(i) - a(i)}{a(i)} \geq 0 \Leftrightarrow  \nonumber \\
&\frac{b^\prime(i)}{a^\prime(i)} - \frac{b(i)}{a(i)} \geq 0. \nonumber 
\end{align}
This follows from (\ref{eqaprime}).

\noindent
\textbf{Case II:}  (\ref{maineqcont}) amounts to
$$
2-\frac{a^\prime(i)}{b^\prime(i)} - \frac{b(i)}{a(i)} \geq 0,
$$
and due to (\ref{equ2}) both $\frac{a^\prime(i)}{b^\prime(i)}$ and $\frac{b(i)}{a(i)}$ are $\leq 1$.

\noindent \textbf{Case III:} $b(i) \geq a(i)$ and $a^\prime(i) \geq b^\prime(i)$ imply $a^\prime(i) \geq b^\prime(i) \geq b(i) \geq a(i)$, which is only compatible with \eqref{eqaprime} if they are all equal, and $s_i^\prime=s_i(\mathcal{C}, d)$.\\

\noindent \textbf{Case IV:} (\ref{maineqcont}) amounts to
\begin{align} 
&\frac{b^\prime(i) - a^\prime(i)}{b^\prime(i)} - \frac{b(i) - a(i)}{b(i)} \geq 0 \Leftrightarrow \nonumber\\
& \frac{a(i)}{b(i)} - \frac{a^\prime(i)}{b^\prime(i)}\geq 0.  \nonumber 
\end{align}
This follows from (\ref{eqaprime}).
\begin{theorem}\label{therorichness} 
The ASW is a rich CQM.
\end{theorem}
{\bf Proof:}
Consider every possible non-trivial clustering $\mathcal{C}$ and construct a distance function $d$ for it such that no other clustering $\mathcal{C}^\prime$ is as good or better in terms of the ASW. 

{\bf Case I:} All clusters in $\mathcal{C}$ contain more than one object. 
Construct $d$ by setting $d(x_i, x_i) = 0$, $d(x_i, x_j) = 1$ if $x_i \sim_{\mathcal{C}} x_j,\  i \neq j$, and $d(x_i, x_j) = 2$  if $x_i \not\sim_{\mathcal{C}} x_j$ for all $ i, j \in X$. Then for all $i\in\mathds{N}_n:$ 
$$
a(i)=1,\ b(i)=2,\ s_i(\mathcal{C}, d) =  0.5,\ \bar{S}(\mathcal{C}, d) = 0.5.
$$
Consider any other clustering $\mathcal{C}^\prime$ (with quantities in the definition of the ASW denoted as $a^\prime(i), b^\prime(i)$). As long as $\mathcal{C}^\prime$ contains no one-point clusters, either in $\mathcal{C}^\prime$ there are $i\neq j$ with $x_i \sim_{\mathcal{C}} x_j$ and $d(x_i, x_j) = 2$, or $i\neq j$ with $x_i \not\sim_{\mathcal{C}} x_j$ and $d(x_i, x_j) = 1$, or both. This results in 
$ a^\prime(i)\geq 1$, $b^\prime(i)\leq 2$, and $s_i(\mathcal{C}^\prime,d)\leq 0.5$
with strict inequality for at least one $i$, therefore $\bar{S}(\mathcal{C}^\prime, d) < 0.5$. One-point clusters $\{x_i\}\in \mathcal{C}^\prime$ yield 
$s_i(\mathcal{C}^\prime,d)=0$ by definition, and no other $s_j(\mathcal{C}^\prime,d)$ can be larger than 0.5, hence again $\bar{S}(\mathcal{C}^\prime, d) < 0.5$, showing that all $\bar{S}(\mathcal{C}^\prime,d)<\bar{S}(\mathcal{C},d)$ for all
$\mathcal{C}^\prime\neq \mathcal{C}$.

{\bf Case II:} $\mathcal{C}$ contains $t$ one-point clusters, $n>t>0$, w.l.o.g, $\{x_1\},\ldots,\{x_t\}$. 
Construct $d$ by setting $d(x_i, x_i) = 0$, $d(x_i, x_j) = 1$ if $x_i \sim_{\mathcal{C}} x_j,\  i \neq j$, $d(x_i, x_j) = 2$  if $x_i \not\sim_{\mathcal{C}} x_j$, and at least one of $i$ and $j$ is larger than $t$, otherwise $d(x_i, x_j) = 2+\frac{1}{2n^2}$. 

Observe for $i=1,\ldots,t:\ s_i(\mathcal{C}, d)=0$, for $i=t+1,\ldots,n:\ s_i(\mathcal{C}, d)=0.5$, therefore $\bar{S}(\mathcal{C}, d) = \frac{(n-t)0.5}{n}$. 

For $t=1$, if $x_t\sim_{\mathcal{C}^\prime}x_i$ for any $i\neq t$, then $s_1(\mathcal{C}^\prime, d)=0$ because  $b^\prime(1) = a^\prime(1) = 2$, and $s_1(\mathcal{C}^\prime, d)=0$ also otherwise. Any difference between $\mathcal{C}$ and $\mathcal{C}^\prime$ will make at least for one $i>1:\ s_i(\mathcal{C}^\prime, d)<0.5$, because for at least one $i>1:\ x_i\sim_{\mathcal{C}^\prime}x_j$ where $x_i \not\sim_{\mathcal{C}} x_j$, leading to $a^\prime(i)>1, b^\prime(i)\le 2$, or $\{x_i\}\in\mathcal{C}^\prime$ with $s_i(\mathcal{C}^\prime, d)=0$, and for all $i:\ s_i(\mathcal{C}^\prime, d)\le 0.5$, so 
$$\bar{S}(\mathcal{C}^\prime, d)<\frac{(n-1)0.5}{n}=\bar{S}(\mathcal{C}, d).$$

If $t\ge 2$, consider first the case that there is no pair $(i,j)$ where  $i\le t$ and $j>t$ with $x_i \sim_{\mathcal{C}^\prime}x_j$. Because $\mathcal{C}\neq \mathcal{C}^\prime$, there must be either $(i,j)$ with both $i,j\le t$ and $x_i \sim_{\mathcal{C}^\prime}x_j$, or $(i,j)$ with both $i,j> t$, $x_i \not\sim_{\mathcal{C}}x_j$, and $x_i \sim_{\mathcal{C}^\prime}x_j$, or $x_i \sim_{\mathcal{C}}x_j$, and $x_i \not\sim_{\mathcal{C}^\prime}x_j$. In all these cases, for $i>t$ always $s_i(\mathcal{C}^\prime, d)\le 0.5$ as before, and for $i\le t:\ a^\prime(i)\ge 2, b^\prime(i)\le 2$, therefore $s_i(\mathcal{C}^\prime, d)\le 0$. Therefore still  $\bar{S}(\mathcal{C}^\prime, d)\le\frac{(n-t)0.5}{n}$. 

Now show ``$<$'', which is required to make $\mathcal{C}$ the unique maximiser of $\bar{S}$.   
If there exists $(i,j)$ with both $i,j\le t$ and $x_i \sim_{\mathcal{C}^\prime}x_j$, $s_i(\mathcal{C}^\prime, d)<0$ because $a^\prime(i)> 2, b^\prime(i)\le 2$, thus $\bar{S}(\mathcal{C}^\prime, d)<\frac{(n-t)0.5}{n}$. If there exists $(i,j)$ with both $i,j> t$, $x_i \not\sim_{\mathcal{C}}x_j$, and $x_i \sim_{\mathcal{C}^\prime}x_j$, $a^\prime(i)> 1, b^\prime(i)\le 2$, therefore $s_i(\mathcal{C}^\prime, d)<0.5$ and again $\bar{S}(\mathcal{C}^\prime, d)<\frac{(n-t)0.5}{n}$. If there exists $(i,j)$ with both $i,j> t$, $x_i \sim_{\mathcal{C}}x_j$, and $x_i \not\sim_{\mathcal{C}^\prime}x_j$, then $a^\prime(i)\ge 1, b^\prime(i)< 2$, therefore $s_i(\mathcal{C}^\prime, d)<0.5$ and again $\bar{S}(\mathcal{C}^\prime, d)<\frac{(n-t)0.5}{n}$. 

The last situation to consider is $t\ge 2$ where there exists $(i,j)$ where  $i\le t$ and $j>t$ with $x_i \sim_{\mathcal{C}^\prime}x_j$. Then 
$$2+\frac{1}{2n^2}\ge a^\prime(i)\ge 2, 2+\frac{1}{2n^2}\ge b^\prime(i)\ge 2,$$ 
so that $s_i(\mathcal{C}^\prime, d)\le \frac{1}{2n^2}$, and this holds for all $i\le t$ (in fact, $s_i(\mathcal{C}^\prime, d)\le 0$ as before unless $x_i\sim_{\mathcal{C}^\prime} x_m$ with any $x_m,\ m\ge t$). Therefore, $\sum_{i\le t}s_i(\mathcal{C}^\prime, d)\le \frac{t}{2n^2}$. Furthermore, let $n^\prime_j=|C|$, where $x_j\in C\in \mathcal{C}^\prime$. With that, 
$$2\ge a^\prime(j)\ge \frac{2+(n^\prime_j-2)}{n^\prime_j-1}=1+\frac{1}{n^\prime_j-1},$$ 
$b^\prime(j)\le 2$, so that 
$$s_j(\mathcal{C}^\prime, d)\le \frac{2-(1+\frac{1}{n^\prime_j-1})}{2}=0.5-\frac{1}{2(n^\prime_j-1)}.$$ 
For $j>t$ so that there is not any $m\le t$ with $x_j \sim_{\mathcal{C}^\prime}x_m$, still $s_j(\mathcal{C}^\prime, d)\le 0.5$. Therefore, 
$$\sum_{i=1}^n s_i(\mathcal{C}^\prime, d)\le 0.5-\frac{1}{2(n^\prime_j-1)}+(n-t-1)0.5+ \frac{t}{2n^2}<(n-t)0.5$$ 
because $\frac{t}{2n^2}<\frac{1}{2(n^\prime_j-1)}$. Overall $\bar{S}(\mathcal{C}^\prime, d)<\frac{(n-t)0.5}{n}$, finishing the proof.  

\section{Simulation study} \label{ssimulation}

We have run a comprehensive simulation study comparing OSil, FOSil, and PAMSil with a number of well established clustering methods from the literature. 
We simulated from a variety of data generating processes (DGPs) with different characteristics that are listed in Table \ref{dgpdef} and illustrated  in Figure \ref{plotoddatasets}. A detailed description is givem in the appendix. 500 data sets were simulated from every DGP. The case of the number of clusters fixed at the true number is investigated as well as the case of an estimated number of clusters. We consider the achieved values of the ASW (the optimisation of which can be of interest in its own right) and the recovery of the ``true'' clusters using the Adjusted Rand Index (ARI; \citet{hubert1985comparing}).

The involved clustering methods are: $k$-means (\citet{lloyd1982least}, \citet{hartigan1979algorithm}, \textit{kmeans}-function in R using \verb|nstart=100| to stabilize the results; apart from this default parameters were used everywhere), $k$-medoids/Partitioning Around Medoids (PAM; \citet{kaufman1987clustering}, \textit{pam}-function in R-package \textit{cluster}; \citet{packagecluster}), average and single linkage (\citet{sokal1958statistical}), Ward's method (\citet{ward1963hierarchical}, all three incorporated in function \textit{agnes} in R-package \textit{cluster}), spectral clustering (algorithm by \citet{ng2001spectral} implemented as \textit{specc} in R-package \textit{kernlab}; \citet{zeileis2004kernlab}), and Gaussian mixture model-based clustering (\citet{fraley1998many}), using function \textit{Mclust} in R-package \textit{mclust}; \citet{fraleymclust}). For PAMSil we have used the standalone C code written by the authors \citet{van2003new}. The function \textit{silhouette} from R-package \textit{cluster} was used for computing ASW-values outside OSil.
\begin{figure}[H]
\begin{center}
\subfloat[DGP 1 ($k=2$)]{
\includegraphics[width=0.33\textwidth]{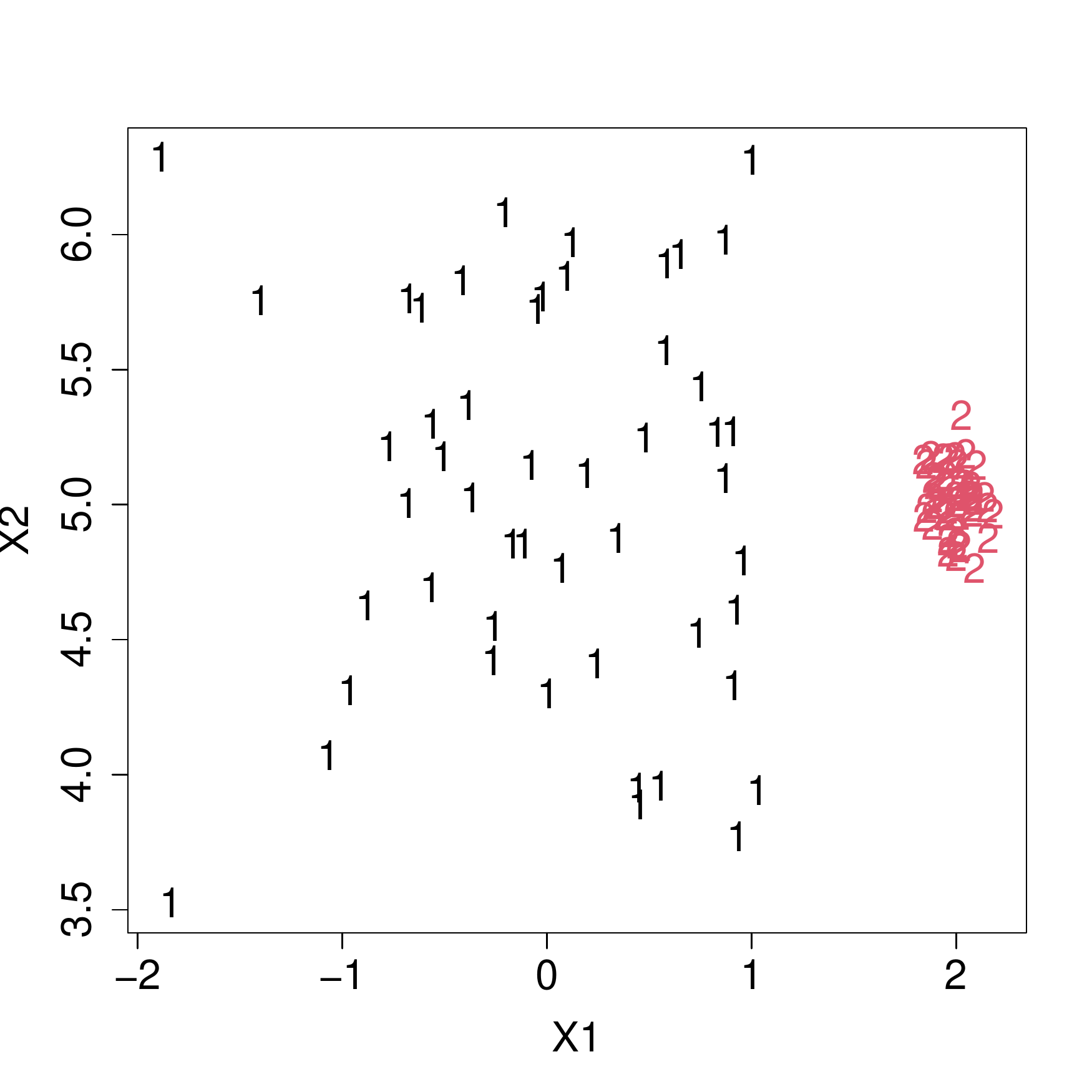} 
}
\subfloat[DGP 2 ($k=3$)]{
  \includegraphics[width=0.33\textwidth]{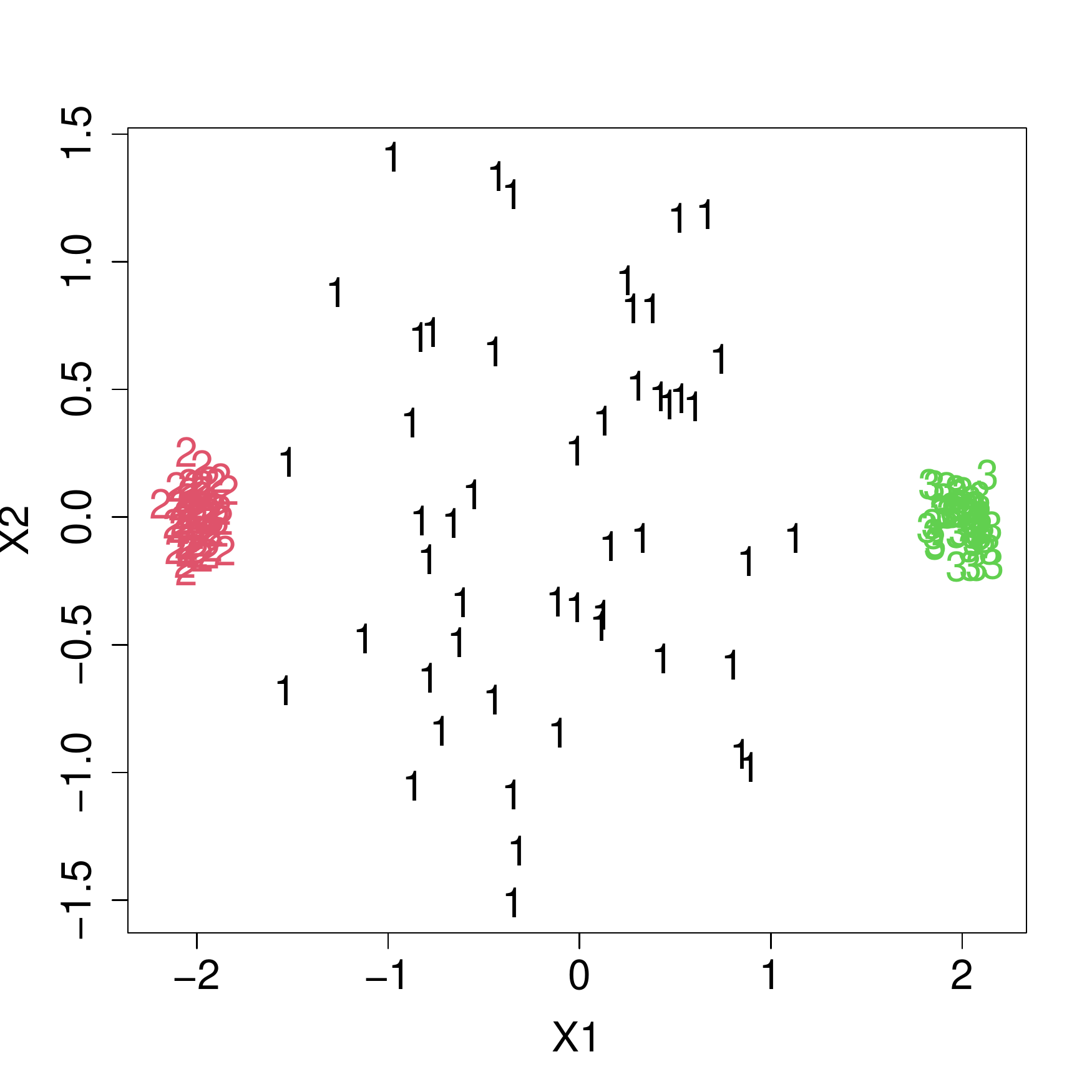}
}
\newline
\subfloat[DGP 3 ($k=4$)]{
  \includegraphics[width=0.33\textwidth]{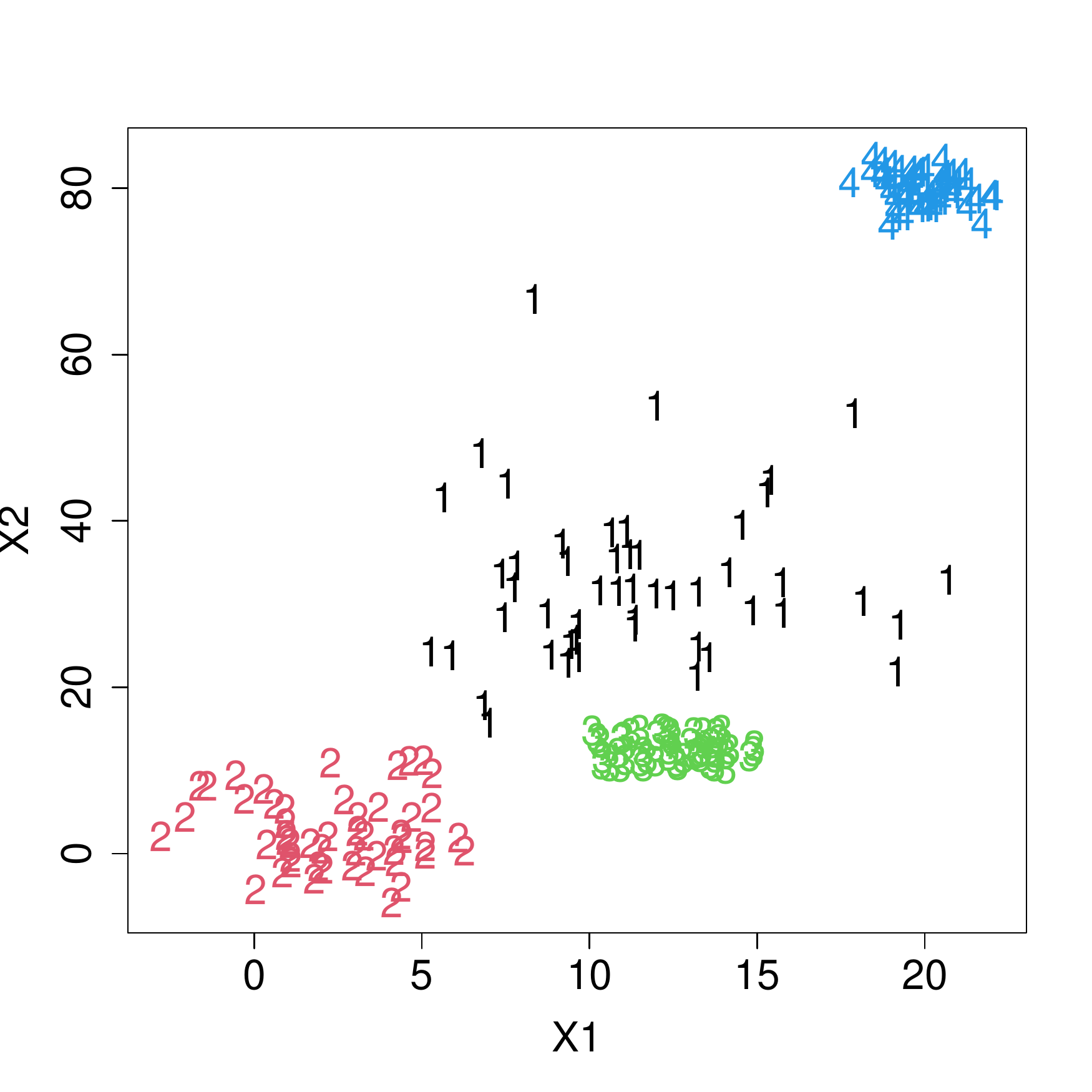}
}
\subfloat[DGP 4 ($k=5$)]{
  \includegraphics[width=0.33\textwidth]{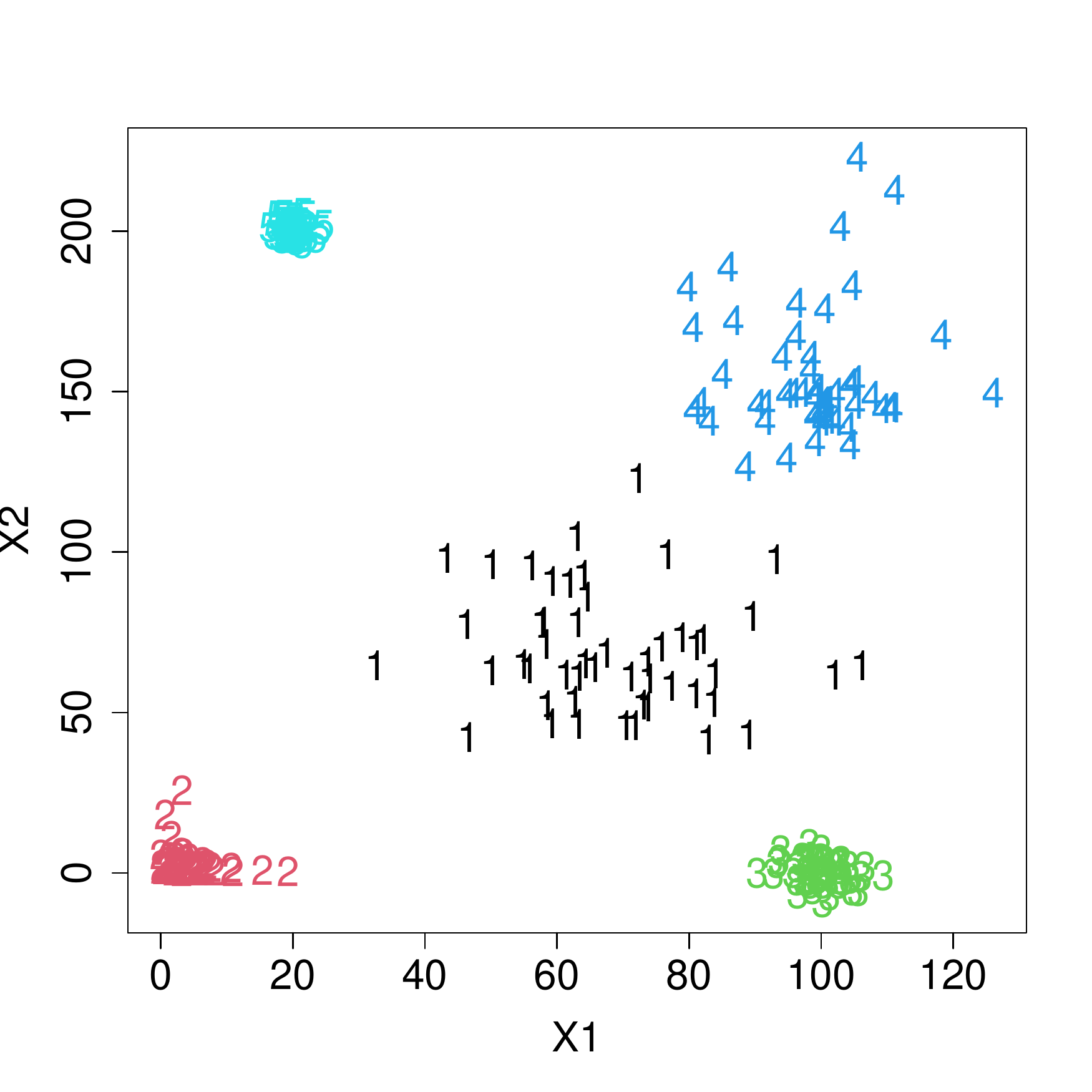}
}
\newline
\subfloat[DGP 5 ($k=6$)]{
  \includegraphics[width=0.33\textwidth]{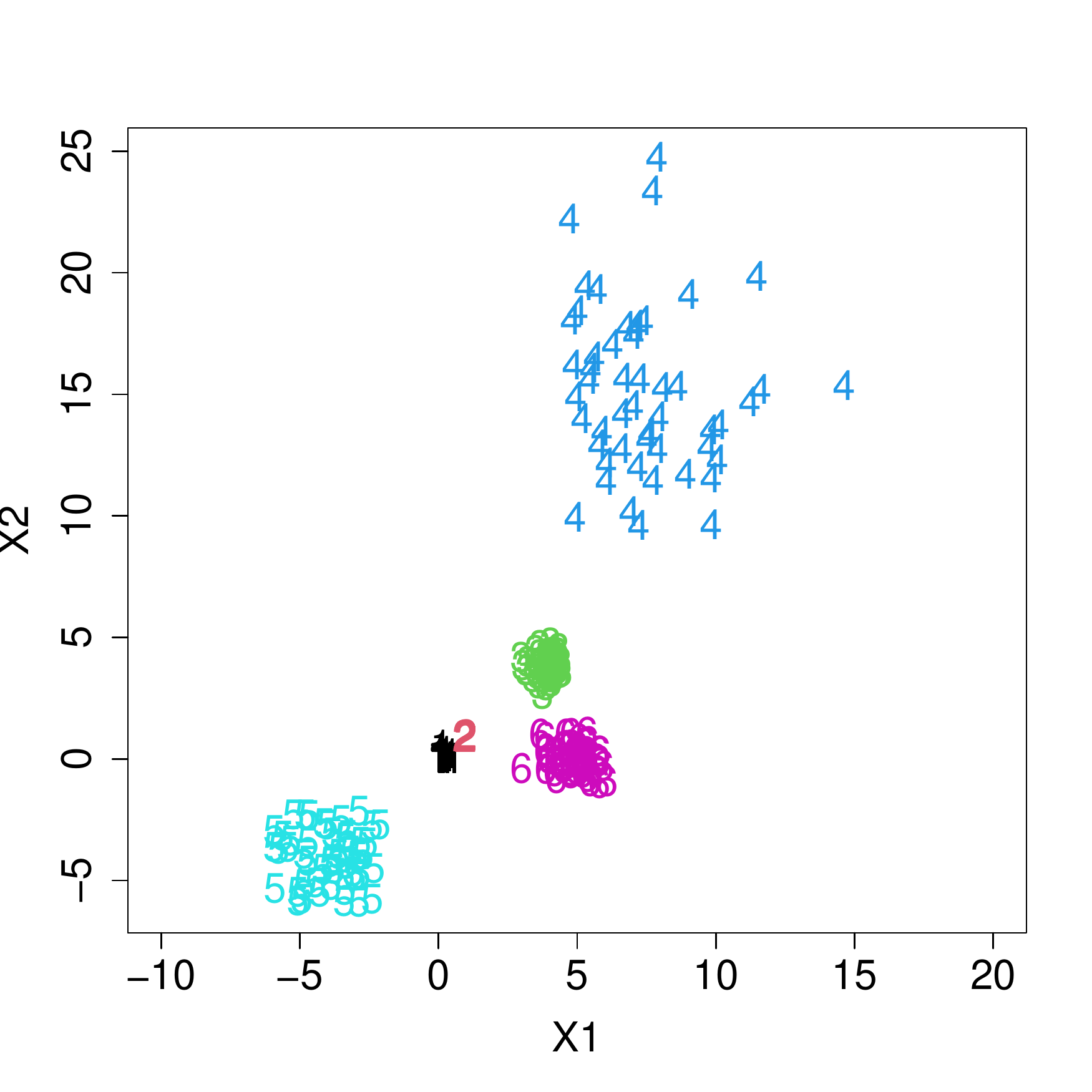}
}
 \subfloat[DGP 6 ($k=6$)]{
  \includegraphics[width=0.33\textwidth]{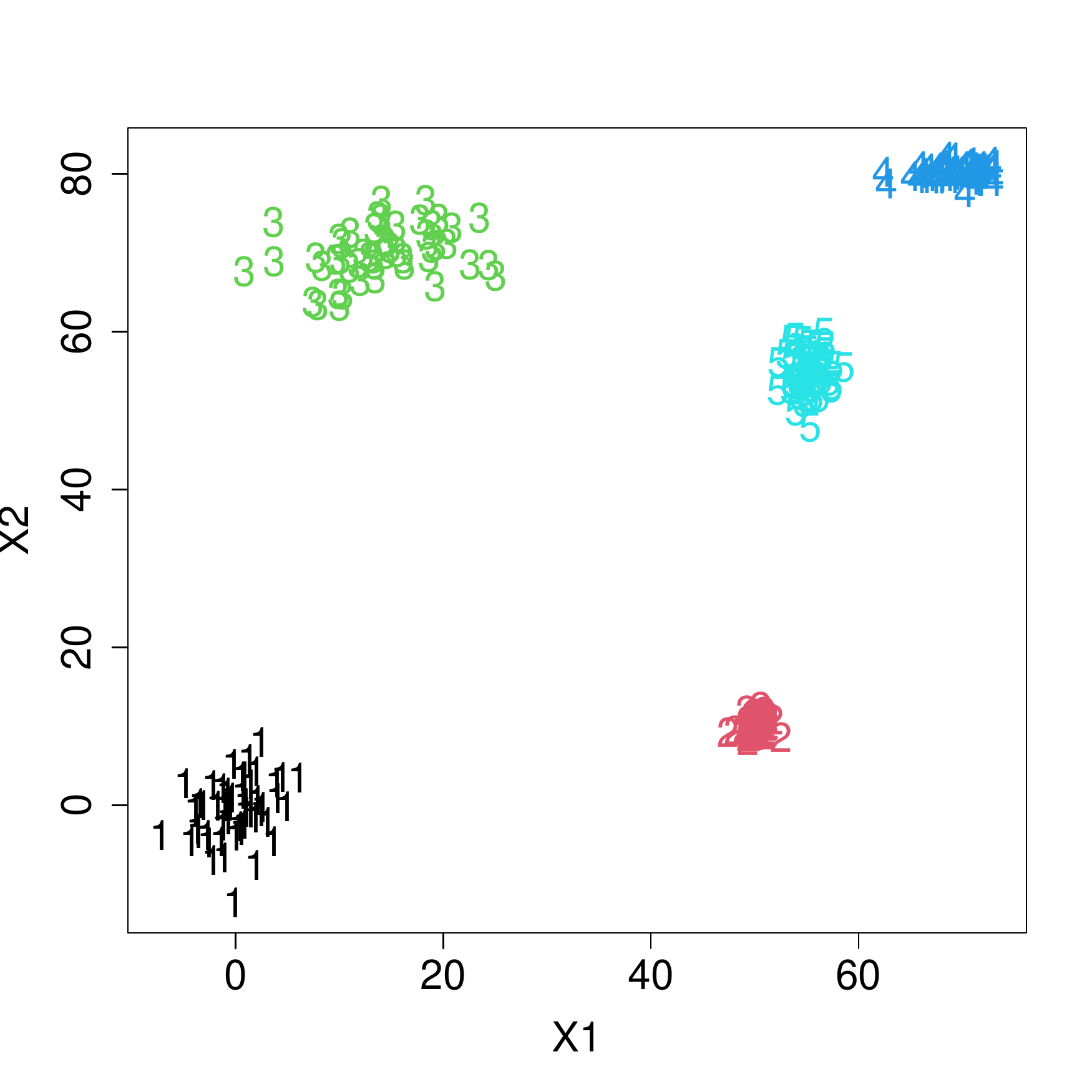}
}
\subfloat[DGP 7 ($k=10$)]{
  \includegraphics[width=0.33\textwidth]{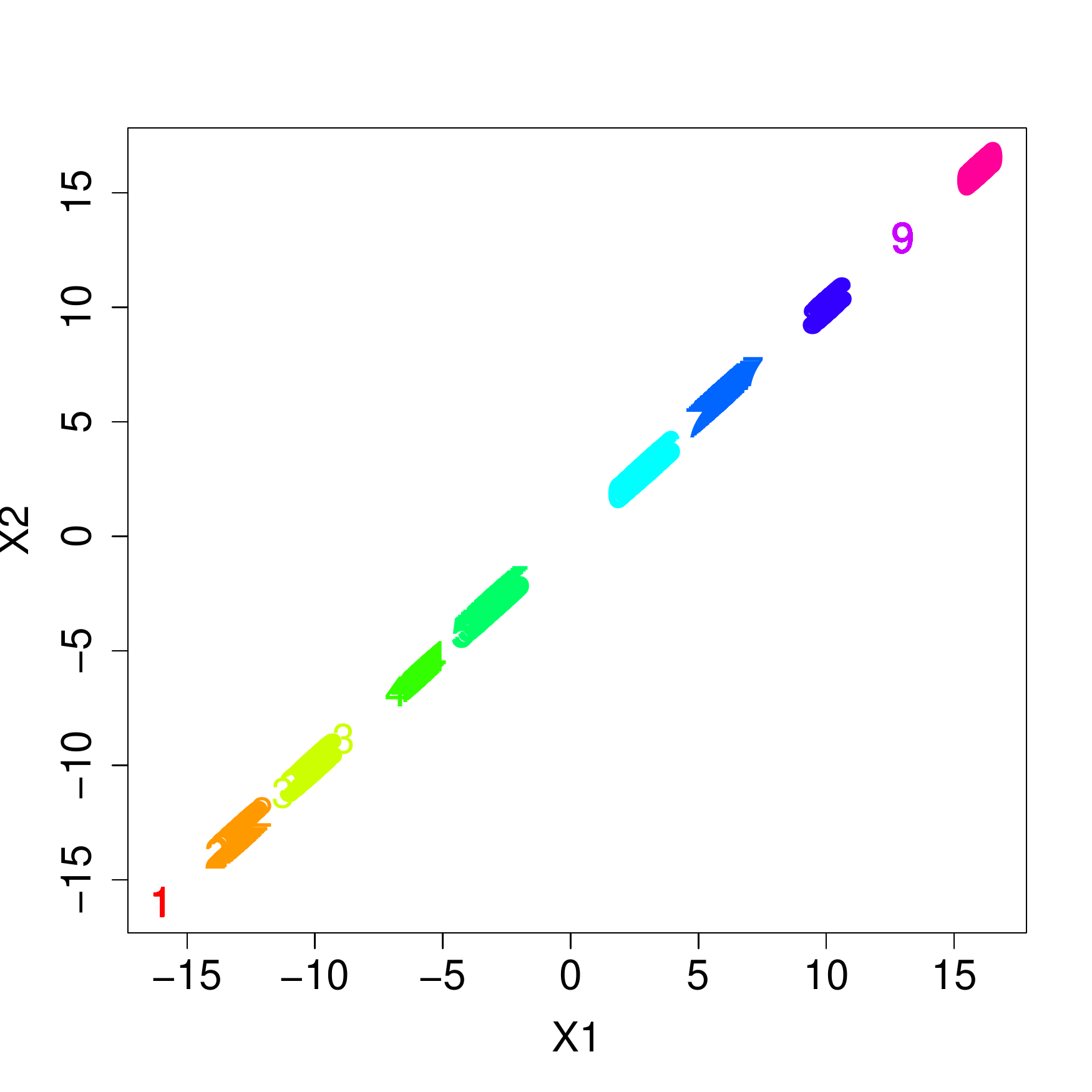}
}
\end{center}
\caption{Plots of the first two dimensions of exemplary data sets 
generated by the DGPs 1-7 with true clusters indicated by colors.}
\label{plotoddatasets}
\end{figure}

\begin{table}[H]
\centering 
\caption{DGPs used in the simulation study. $k$: number of clusters, $p$: number of dimensions, $n$:  number of observations. See the Appendix for precise descriptions.} 
\fontsize{11}{11}\selectfont
\begin{threeparttable}
\begin{tabular}{ l l l l l l} 
\toprule 
DGP & $k$ & $p$ & Distributions & Cluster size & $n$\\
\midrule
\textbf{DGP1 } & 2 & 2 & Spherical Gaussians  & 50 & 100\\
\textbf{DGP2}  & 3 & 2 & Spherical Gaussians  & 50 & 150\\
\textbf{DGP3 } & 4 & 2 & $t$, uniform and two Gaussians & 50 & 200\\
\textbf{DGP4}  & 5 & 2 & $F$, $\chi^2$, $t$, skew Gaussian, Gaussian & 50 & 250\\
\textbf{DGP5}  & 6 & 2 & six different distributional shapes & 50 & 300\\
\textbf{DGP6}  & 5 & 5 & five Gaussians with different & 50 & 250\\
& & & within-cluster dependence structures & \\
\textbf{DGP7} & 10&500& 10-cluster structure on 1-d hyperplane  & 50 & 500\\
&&& in 500-d space. &&\\
\textbf{DGP8} & 7 & 60 & Experiment of \cite{van2003new}  & varying  & 500  \\
&&& simulated genes defining 3 patient groups &&\\ 
\textbf{DGP9} & 3 & 1000 & First 100 dimensions informative  & 40 & 120\\
&&& spherical Gaussian &&\\ 
\bottomrule
\end{tabular}
      \end{threeparttable}
      \label{dgpdef}
\end{table}

Estimating $k$ was done by optimizing the ASW with two exceptions. Gaussian mixture model-based clustering was used with the BIC, which is the standard choice for mixtures. Exemplary for many other possible combinations of clustering method and method to estimate the number of clusters, we also ran $k$-means with the gap statistic (\cite{tibshirani2001estimating}) as implemented in the R-function \textit{clusGap}.

\subsection{Results and discussion}
Results of the simulations for the DGPs 1-3 are in Table \ref{tab:OFOSIL1-3}; for the DGPs 4-6 in Table \ref{tab:OFOSIL4-6}, and for DGPs 7-9 in Table \ref{tab:OFOSIL7-9}. These tables report average ARI and ASW and their estimated standard errors over the simulation runs for both fixed and estimated $k$ as well as the percentage of runs out of those for estimated $k$ in which the true $k$ was estimated.

\begin{table}[H] \centering 
  \caption{Simulation results for DGPs 1-3. PPR is the percentage of runs in which the true $k$ was estimated.} 
\fontsize{10}{10}\selectfont
\begin{tabular}{@{\extracolsep{-.1pt}}lcccccccccccc} 
\hline 
\hline 
&  \multicolumn{4}{c}{Fixed $k$}   & \multicolumn{5}{c}{Estimated $k$}  &  \\ 
\cline{2-10}
Methods & ASW & SE & ARI & SE  & ASW & SE & ARI & SE  & PPR  \\
\hline
\textbf{DGP1} \\
k-means &   0.665 & 0.001 & 0.808 & 0.004 & 0.665 & 0.001 & 0.796 & 0.004 & 92\\
gap-k-means & & & & & 0.665 & 0.001  & 0.760 & 0.005 & 70\\ 
PAM &  0.665 & 0.001 & 0.834 & 0.003 & 0.666 & 0.001 & 0.816 & 0.004 & 92  \\ 
average &   0.622 & 0.004 & 0.687 & 0.016 & 0.655 & 0.001 & 0.812 & 0.007 & 72 \\ 
Ward's & 0.654 & 0.001 & 0.936 & 0.004 & 0.656 & 0.001 & 0.891 & 0.006 & 86  \\ 
Single & 0.410 & 0.007 & 0.167 & 0.016 & 0.545 & 0.005 & 0.785 & 0.013 & 29\\
BIC-mixture &  0.646 & 0.001 & 0.993 & 0.001 &   0.643 & 0.001 & 0.991 & 0.001 & 99  \\ 
spectral &  0.643 & 0.003 & 0.923 & 0.005 & 0.656 &  0.001 & 0.906 & 0.009 & 89 \\ 
PAMSil & 0.667 &  0.001 & 0.848 & 0.006 &  0.668 & 0.001 & 0.816 & 0.008 & 87   \\ 
OSil &   0.666 & 0.001 & 0.847 & 0.004 & 0.667 & 0.001 & 0.812 & 0.005 & 86\\
FOSil  &   0.659 & 0.001 & 0.785 & 0.007 & 0.662 & 0.001 & 0.803 & 0.005 & 94 \\
\textbf{DGP2} \\
k-means & 0.711 & 0.001 & 0.844 & 0.004 & 0.719 & 0.001 & 0.805 & 0.005 & 39 \\ 
gap-k-means & & & & & 0.714 & 0.001  & 0.801 & 0.002 & 39\\ 
pam & 0.711 & 0.001 & 0.85 & 0.003 & 0.719 & 0.001 & 0.809 & 0.004 & 38 \\ 
average & 0.671 & 0.004 & 0.773 & 0.013 & 0.711 & 0.001 & 0.821 & 0.005 & 26 \\ 
Ward's & 0.696 & 0.002 & 0.934 & 0.005 & 0.708 & 0.001 & 0.844 & 0.006 & 30 \\ 
single & 0.348 & 0.021 & 0.404 & 0.023 & 0.611 & 0.008 & 0.818 & 0.014 & 11 \\ 
BIC-mixture &  0.652 & 0.002 & 0.863 & 0.002 & 0.682 & 0.002 & 0.991 & 0.001 & 91 \\
spectral & 0.628 & 0.013 & 0.898 & 0.012 & 0.700 & 0.002 & 0.877 & 0.006 & 50 \\ 
PAMSil & 0.710 & 0.001 & 0.859 & 0.004 & 0.721 & 0.001 & 0.804 & 0.004 & 22 \\ 
OSil & 0.712 & 0.001 & 0.856 & 0.004 & 0.722 & 0.001 & 0.806 & 0.004 & 25 \\ 
FOSIL & 0.705 & 0.002 & 0.67 & 0.015 & 0.714 & 0.001 & 0.815 & 0.006 & 48 \\ 
\textbf{DGP3} \\
k-means & 0.674 & 0.002 & 0.815 & 0.005 & 0.764 & 0.001 & 0.318 & 0.001 & 0.2 \\ 
gap-k-means & & & & & 0.692 & 0.001  & 0.816 & 0.006 & 38\\ 
pam & 0.702 & 0.001 & 0.912 & 0.001 & 0.765 & 0.001 & 0.319 & 0.001 & 0.2 \\ 
average & 0.644 & 0.001 & 0.647 & 0.004 & 0.764 & 0.001 & 0.325 & 0.000 & 0 \\ 
Ward's & 0.689 & 0.001 & 0.96 & 0.003 & 0.762 & 0.001 & 0.328 & 0.001 & 0 \\ 
single & 0.54 & 0.005 & 0.464 & 0.009 & 0.75 & 0.002 & 0.341 & 0.004 & 0.6 \\ 
BIC-mixture & 0.676 & 0.001 & 0.996 & 0.000 & 0.664 & 0.002 & 0.964 & 0.002 & 53 \\
spectral &  0.563 & 0.009  & 0.329 & 0.001 &  0.761 & 0.001 & 0.836 & 0.007 & 0\\ 
PAMSil &  0.703 & 0.001 & 0.913 & 0.002  & 0.768 & 0.001 & 0.320  & 0.001 & 0  \\ 
OSil & 0.703 & 0.001 & 0.91 & 0.003 & 0.768 & 0.001 & 0.322 & 0.000 & 0 \\ 
FOSil & 0.699 & 0.001 & 0.502 & 0.002 & 0.767 & 0.001 & 0.32 & 0.000 & 0 \\ 
\hline
\end{tabular} 
\label{tab:OFOSIL1-3}
\end{table}

\begin{table}[H] \centering 
  \caption{Simulation results for DGPs 4-6. PPR is the percentage of runs in which the true $k$ was estimated.} 
\fontsize{10}{10}\selectfont
\begin{tabular}{@{\extracolsep{-.1pt}}lcccccccccccc} 
\hline 
\hline 
&  \multicolumn{4}{c}{Fixed $k$}   & \multicolumn{5}{c}{Estimated $k$}  &  \\ 
\cline{2-10}
Methods & ASW & SE & ARI & SE  & ASW & SE & ARI & SE  & PPR  \\
\hline
\textbf{DGP4}\\
k-means & 0.727 & 0.006 & 0.887 & 0.007 & 0.791 & 0.002 & 0.962 & 0.003 & 65 \\ 
gap-k-means & & & & & 0.728 & 0.004  & 0.814 & 0.010 & 47\\ 
pam & 0.818 & 0.000 & 0.99 & 0.000 & 0.818 & 0.000 & 0.99 & 0.000 & 100 \\ 
average & 0.808 & 0.002 & 0.975 & 0.003 & 0.816 & 0.000 & 0.99 & 0.001 & 93 \\ 
Ward's & 0.817 & 0.000 & 0.992 & 0.001 & 0.817 & 0.000 & 0.992 & 0.001 & 99 \\ 
single & 0.694 & 0.005 & 0.859 & 0.005 & 0.777 & 0.002 & 0.965 & 0.003 & 42 \\ 
BIC-mixture & 0.8 & 0.001 & 0.98 &  0.001 & 0.761 & 0.003 & 0.956 & 0.002 & 37 \\ 
spectral &   0.645 &  0.010 &  0.962 &  0.003 & 0.785 & 0.002 &  0.885 &    0.006 & 58\\ 
PAMSil &  0.818  & 0.000 & 0.993 & 0.000 &  0.818 & 0.000 & 0.993 & 0.000  & 97  \\ 
OSil & 0.818 & 0.000 & 0.993 & 0.000 & 0.818 & 0.000 & 0.993 & 0.000 & 98 \\ 
FOSil & 0.817 & 0.000 & 0.987 & 0.002 & 0.817 & 0.000 & 0.99 & 0.001 & 99\\ 
\textbf{DGP5} \\
k-means & 0.659 & 0.004 & 0.769 & 0.007 & 0.723 & 0.001 & 0.745 & 0.013 & 35 \\ 
gap-k-means & & & & & 0.626 & 0.005  & 0.565 & 0.009 & 2\\ 
pam & 0.743 & 0.001 & 0.957 & 0.003 & 0.745 & 0.000 & 0.975 & 0.001 & 82 \\ 
average & 0.581 & 0.001 & 0.342 & 0.006 & 0.700 & 0.000 & 0.232 & 0.010 & 0 \\ 
Ward's & 0.717 & 0.000 & 0.775 & 0.000 & 0.726 & 0.001 & 0.766 & 0.005 & 48 \\ 
single & 0.572 & 0.002 & 0.581 & 0.007 & 0.68 & 0.001 & 0.191 & 0.008 & 2 \\ 
BIC-mixture & 0.696 & 0.001 & 0.778 & 0.001  & 0.731 & 0.001 & 0.821 & 0.001 & 4 \\ 
spectral & 0.646 & 0.006 & 0.766 & 0.005 & 0.725 & 0.001 & 0.798 & 0.012 & 33 \\ 
PAMSil & 0.748 & 0.000 & 0.995 & 0.001 & 0.748 & 0.000 & 0.995 & 0.001 & 95 \\ 
OSil & 0.747 & 0.001 & 0.974 & 0.003 & 0.748 & 0.000 & 0.988 & 0.001 & 87 \\ 
FOSIL & 0.747 & 0.001 & 0.785 & 0.005 & 0.748 & 0.000 & 0.815 & 0.001 & 92 \\ 
\textbf{DGP6} \\
k-means &   0.649 & 0.007 & 0.739 & 0.009 & 0.77 & 0.003 & 0.852 & 0.008 & 28\\
gap-k-means & & & & & 0.773 & 0.002  & 0.589 & 0.008 & 7\\ 
PAM &    0.865 & 0 & 1 & 0 & 0.865 & 0 & 1 & 0 & 100    \\ 
average &    0.865 & 0 & 1 & 0 & 0.865 & 0 & 1 & 0 & 100 \\ 
Ward's & 0.865 & 0 & 1 & 0 & 0.865 & 0 & 1 & 0 & 100   \\ 
Single & 0.865 & 0 & 1 & 0 & 0.865 & 0 & 1 & 0 & 100\\
BIC-mixture &  0.865 & 0 & 1 & 0 & 0.865 & 0 & 1 & 0 & 100  \\ 
spectral &  0.618 &    0.023 & 0.881 &    0.016    &  0.797  &   0.007 &                    0.834 &  0.016 & 43  \\ 
PAMSil &    0.865 & 0 & 1 & 0 & 0.865 & 0 & 1 & 0 & 100\\ 
OSil &   0.865 & 0 & 1 & 0 & 0.865 & 0 & 1 & 0 & 100 \\
FOSil  &  0.865 & 0 & 0.980 & 0 & 0.865 & 0 & 0.98 & 0 & 100  \\
\hline
\end{tabular} 
\label{tab:OFOSIL4-6}
\end{table}

\begin{table}[H] \centering 
  \caption{Simulation results for DGPs 8-10. PPR is the percentage of runs in which the true $k$ was estimated.} 
  \fontsize{10}{10}\selectfont
\begin{tabular}{@{\extracolsep{-.1pt}}lccccccccccc} 
\hline 
\hline 
&  \multicolumn{4}{c}{Fixed $k$}   & \multicolumn{5}{c}{Estimated $k$}  &  \\ 
\cline{2-10}
Methods & ASW & SE & ARI & SE & ASW & SE & ARI & SE & PPR    \\
\hline
\textbf{DGP7} \\
k-means & 0.754 & 0.003 & 0.767 & 0.004 & 0.825 & 0.002 & 0.863 & 0.004 & 21 \\
gap-k-means & & & & & 0.724 & 0.003  & 0.649 & 0.006 & 3\\ 
PAM &   0.921 & 0.001 & 1 & 0 & 0.921 & 0.001 & 1 & 0 & 100   \\ 
average &  0.921 & 0.001 & 1 & 0 & 0.921 & 0.001 & 1 & 0 & 100 \\ 
Ward's &  0.921 & 0.001 & 1 & 0 & 0.921 & 0.001 & 1 & 0 & 100   \\ 
Single & 0.92 & 0.001 & 1 & 0 & 0.921 & 0.001 & 1 & 0 & 99.6\\
BIC-mixture &  0.921 & 0.001 & 1 & 0 &   0.860 & 0.001 & 0.896 & 0 &    0 \\ 
spectral & 0.584  & 0.031  &  0.798 & 0.017  &  0.830 &  0.011 &  0.854 & 0.032 &  14.6 &   \\ 
PAMSil &  0.921 & 0.001 & 1  &    0 &    0.921 & 0.001 & 1  &    0  &   100   \\ 
OSil &   0.921 & 0.001 & 1 &    0 &    0.921 & 0.001 & 1 &    0 &    100    \\
FOSil  &     0.921 & 0.001 & 1 & 0 &    0.921 & 0.001 & 1 & 0  &    100 \\
\textbf{DGP8}\\
k-means & 0.178 & 0.004 & 0.823 & 0.010 & 0.231 & 0.000 & 0.949 & 0.005 & 47\\            
gap-k-means & & & & & 0.180 & 0.003  & 0.747 & 0.010 & 16\\ 
PAM & 0.104 & 0.003 & 0.528 & 0.009 & 0.118 & 0.003 & 0.591 & 0.008 & 20 \\            
average & 0.242 & 0.000  & 0.999 & 0 & 0.242 & 0.000 & 0.999 & 0 & 100 \\
Ward's & 0.242 & 0.000 & 1 & 0 & 0.242 & 0.000 & 1 & 0 & 100 \\ 
single & 0.186 & 0.001 & 0.703 & 0.007 & 0.217 & 0.000 & 0.522 & 0.016 & 12\\            
BIC-mixture & 0.191  & 0.000 & 0.695 & 0.006 & 0.203 & 0.000 & 0.785 & 0.006 &  24  \\                
spectral & 0.081 & 0.004 & 0.546 & 0.011  & 0.220 & 0.000 & 0.546 & 0.012 & 16   \\              
PAMSil & 0.242 &  0.000 & 1 & 0.000  & 0.243 &  0.000 &  0.907  & 0.013 & 91 \\ 
OSil & 0.242 & 0.000 & 0.999 & 0 & 0.255 & 0.000 & 0.230 & 0.018 & 22 \\ 
FOSIL & 0.240 & 0.000 & 0.996 & 0.005 & 0.247 & 0.000 & 0.609 & 0.018 & 68 \\
\textbf{DGP9}\\
k-means &   0.488 & 0.006 & 0.827 & 0.013 & 0.56 & 0.001 & 0.865 & 0.010 &   69 \\
gap-k-means & & & & & 0.074 & 0.003  & 0.654 & 0.006 & 6\\ 
PAM &  0.573 & 0 & 1 & 0 & 0.573 & 0 & 1 & 0 &   100   \\ 
average &  0.573 & 0 & 1 & 0 & 0.573 & 0 & 1 & 0 &   100  \\ 
Ward's &    0.573 & 0 & 1 & 0 & 0.573 & 0 & 1 & 0 &   100   \\ 
Single &  0.573 & 0 & 1 & 0 & 0.573 & 0 & 1 & 0 &   100\\
BIC-mixture &  0.573 & 0 & 1 & 0 &  0.573 & 0 & 1 & 0 &  100   \\ 
spectral & 0.544  & 0.005  &  0.956 &  0.007 & 0.569  & 0.001 & 0.966 & 0.006 & 92  \\ 
PAMSil &   0.573   &  0 &   1  &  0 & 0.573  & 0  &  1 & 0 &  100 \\ 
OSil &    0.573 & 0 & 1 & 0 & 0.573 & 0 & 1 & 0 &   100 \\
FOSil  &  0.573 & 0 & 1 & 0 & 0.573 & 0 & 1 & 0 &   100\\
\hline
\end{tabular} 
\label{tab:OFOSIL7-9}
\end{table}

As could be expected, the methods for (at least locally) optimizing the ASW
achieve the best ASW values, although sometimes PAM and occasionally  
some other methods find solutions with competitive ASW values. Occasionally,
PAM can find a slightly better ASW value than FOSil, which often loses some
optimization power compared with OSil and PAMSil. OSil yields sometimes higher
ASW values than PAMSil, but sometimes slightly worse; for achieving the 
highest possible ASW value, it is advisable to run both. 

Regarding the recovery of the modelled clusters, OSil and PAMSil clearly do 
the best job for DGP 5 compared to the other methods both for 
estimated and fixed $k$. They are best by a small margin for DGP 4 and 
up with the best for DGPs 6, 7, and 9, and for fixed $k$ only for DGP 8.
FOSil drops a bit in comparison in some of these.
DGP 4 and 5 in particular are difficult for most standard methods because of
the different distributional shapes of the clusters, with variance-covariance
structures also strongly varying. As could be expected, Gaussian mixture 
model-based clustering is best in the setups with all or mostly Gaussian
within-cluster distributions. For DGP 3 and estimated $k$, it 
is the only method that does a good job.
   
The performance of PAMSil and OSil for DGPs 1-3 is surprisingly relatively 
better for fixed $k$ than for estimated $k$,
contrasting with the popularity of the ASW for estimating $k$.  In DGP 8,
where Ward's method and average linkage perform best, PAMSil, OSil, and FoSil
sometimes find solutions with better ASW for wrong values of $k$ (often $k=2$)
when estimating $k$, which diminishes their ARI-performance for estimated $k$. 
This is in line with the skepticism that can be found in 
\cite{hennig15paramboot} 
regarding naively maximizing the ASW for estimating $k$, see also Section
\ref{sfrance}. The fact that OSil can find a better ASW for
another than the modelled $k$ more often than PAMSil, leads to a 
disappointingly low ARI here.

Regarding the other methods and the recovery of the modelled clusters, 
Ward's method performs very well across the board
with exception of DGP 5, where it particularly drops for estimated $k$. It is 
almost always better than $k$-means despite being worse at optimizing the
$k$-means objective function. Probably the hierarchical
structure helps it to adapt better to the varying cluster shapes; 
similarly PAMSil can occasionally do better than OSil even where the value of 
the objective function ASW is worse.

The other methods have some mixed results, with spectral, $k$-means and single
linkage performing mostly clearly behind the top methods. PAM has some drops
in quality but performs generally well and often similar to PAMSil and OSil;
it achieves the best ASW values out of the methods that do not attempt to
optimize it. The results of average linkage are overall with one exception 
on a slightly lower level than PAM. The Gaussian 
mixture drops in quality in DGPs 5 and 8, but is good
in the other setups. The gap statistics with $k$-means achieves a good result 
for DGP 3 but does worse than $k$-means with the ASW otherwise.

Overall PAMSil and OSil do a good job, particularly with the mixed distribution
shapes and fixed $k$, with none of the two clearly superior to the other. Some
more caution is required when estimating $k$. The simulation shows that there
are situations in which these methods are best, and therefore they can be seen
as a valuable addition to the cluster analysis toolbox without generally
outperforming the competition. Particularly in situations with Gaussian 
distributions only, Gaussian mixture-based clustering is to be preferred.
Ward's method presents itself as a good allrounder. The 
computationally less intensive FOSil can sometimes not 
keep up with OSil and PAMSil, but it has some very good results estimating the 
number of clusters, better than OSil and PAMSil in DGPs 1, 2, and 4.  

\subsection{An experiment with outliers}\label{soutliers}
The ASW and OSil do not rely on any distributional assumption, and it is 
therefore
of interest whether they are less affected by outliers than methods that rely 
on a normal distribution assumption or on sums of squares such as $k$-means
or Gaussian mixture model-based clustering. Unfortunately, as shown in 
\cite{HenC08}, as a method for estimating the number of clusters, the ASW is
not perfectly robust. An outlier that is very far from the rest of the data 
can prompt the ASW to select a solution with two clusters, where one cluster 
is just the outlier, and the other cluster is everything else taken together.
The reason for this is that if $k$ is made larger, and less
strongly separated data subsets that intuitively still qualify as ``clusters''
are separated, the corresponding $b(i)$-values in the ASW are much smaller than
if these subsets are put together and there is another cluster far away from 
them, which then is the closest (for $k=2$ in fact the only) other cluster. 

The ASW can however add some robustness in case that the outliers are not 
that extreme. We generated two more data sets 
from DGP 6, one with 250 and one with 1000 observations, for which most methods
did a very good job, see Table \ref{tab:OFOSIL4-6}. We added
outliers, one at $(70,70,30,0,55)$ in some distance from the clusters but not 
outside the original data range in any variable, and another one at 
$(-100,-100,-100,-100,-100)$, outlying on all variables. Four datasets, namely 
with 250 and 100 original points, and with the first only 
or two outliers are considered. The original 
clustering pattern is very clear (see Figure \ref{plotoddatasets}), and it would
be desirable to still find this structure with outliers present. 
OSil with 
estimated $k$ isolates each outlier in a one-point cluster, leaving the original
clustering intact, which seems sensible. Standard 
hierarchical methods do the same if the ASW is used to decide the number of 
clusters. FOSil needs the outliers in the random subsample used for initial 
clustering to arrive at the same result. Otherwise it assigns the outliers 
to the nearest original cluster, still
leaving the original structure intact, but causing some very large 
within-cluster distances. The same happens with Gaussian model-based 
clustering for three of the 
four data sets. With 1000 observations and the
first outlier only, the first outlier is merged with a small part of an original cluster, which is split up. $k$-means splits original clusterings in the 
presence of outliers in its
solutions with $k=5, 6,$ and 7; the gap statistic chooses $k=5$ anyway with one
outlier, but $k=1$ with two outliers.
PAM and spectral clustering also avoid one-point-clusters and split original
clusters and join parts with outliers if present. In these 
examples, the ASW both for choosing $k$ and used by OSil and FOSil for given 
$k$ treats the outliers better than competing methods, due to its ability
to isolate well separated one-point (or very small) clusters. 
 
\section{Applications}\label{sappl}
Section \ref{sscrna} is devoted to four genetic data sets using the 
Euclidean distance, which come with ground truth information.
Sections \ref{sveronica} and \ref{sfrance} explore the application of OSil, 
FOSil, and PAMSIL to 
data represented by other dissimilarity measures than the Euclidean distance.
\subsection{Clustering single cell RNA sequencing data} \label{sscrna}
Clustering of single cell RNA sequencing (scRNA-seq) data is a vital field. It is of interest in its own respect, and it can be used as first step for further analysis. Since much of the downstream analysis is based on clustering, the final conclusions may be strongly affected by it. Definition or discovery of new cell types via clustering is an important area of research in the field.  Many different studies have been conducted on various organs either during development or at fixed time to discover several new putative cell sub-populations using novel clusters, for instance, in early embryonic development (\citet{biase2014cell}, \citet{goolam2016heterogeneity}) or various regions of the brain (\citet{zeisel2015cell}).  We consider scRNA-seq data clustering by the proposed methods for a number of published  data sets for which the true cell types were originally identified by the authors. 

The data sets are listed in Table \ref{realdata}. We 
followed \cite{lun2016step} normalizing them.
All data sets were represented by principal components (PCs). Euclidean distances on these were used for clustering. scRNA-seq data are typically of low sample size and high dimensionality, and dimension reduction is routinely applied. The number of principal components was generally chosen as $q$ so that from $q$ to $q+1$ PCs there was still a substantial drop in explained variance per PC, but from $q+1$ PCs onward every PC would only account for low percentages of the variance with little further drop from one PC to the next. For $j>0$, variance PC $q+j$ was only bigger than variance PC $q+j+1$ by substantially less than factor 2.
The maximum number allowed for the estimation of  number of clusters was 12.  

 \begin{table}[H]
 \centering 
\caption{scRNA-seq data sets} 
\fontsize{11}{11}\selectfont
\begin{tabular}{ l c c c c c c} 
\toprule 
Data & $n$ & number of genes & number of PCs & $k$   \\
\midrule
\citet{yan2013single}  &  90 &  20214 & 2 & 7 \\ 
\citet{biase2014cell}  & 49 & 25737 & 3 & 3\\
\citet{goolam2016heterogeneity} & 124 & 41324 & 4 & 5 \\
\citet{kolodziejczyk2015single} & 704 & 38577 & 5 & 3  \\
\bottomrule
\label{realdata}
\end{tabular}
\end{table} 

\citet{yan2013single} is a study  of human embryonic development. The authors identified 7 cell types (development stages) as oocyte (3 samples), zygote (3 samples), 2-cell (6 samples),  4-cell (12 samples), 8-cell (20 samples), lateblast (30 samples), and morula (16 samples). The first two principal components were used, which explain 80\% of the variance.

\citet{biase2014cell}  studied the cell fate decision during early embryo development. There are  1-cell (9 samples), 2-cell (20 samples), and 4-cell (20 samples) embryos. The first three principal components were used. They explain 43\% of the variance.

\citet{goolam2016heterogeneity} studied pre-implantation development.  There are five distinct cell types, 2-cell (16 samples), 4-cell (64 samples), 8-cell (32 samples), 16-cell (6 samples) and 32-cell (6 samples). The first four principal components were used, which explain 60\% of the variance.  

\citet{kolodziejczyk2015single} studied mouse embryonic stem cell growth under different culture conditions.  The three culture conditions are serum (250 cells), 2i (295 cells) and 2ai (159 cells). The first five principal components were used, which explain 36\% of the variance.

\begin{figure}[H]
\centering
  \includegraphics[width=0.48\textwidth]{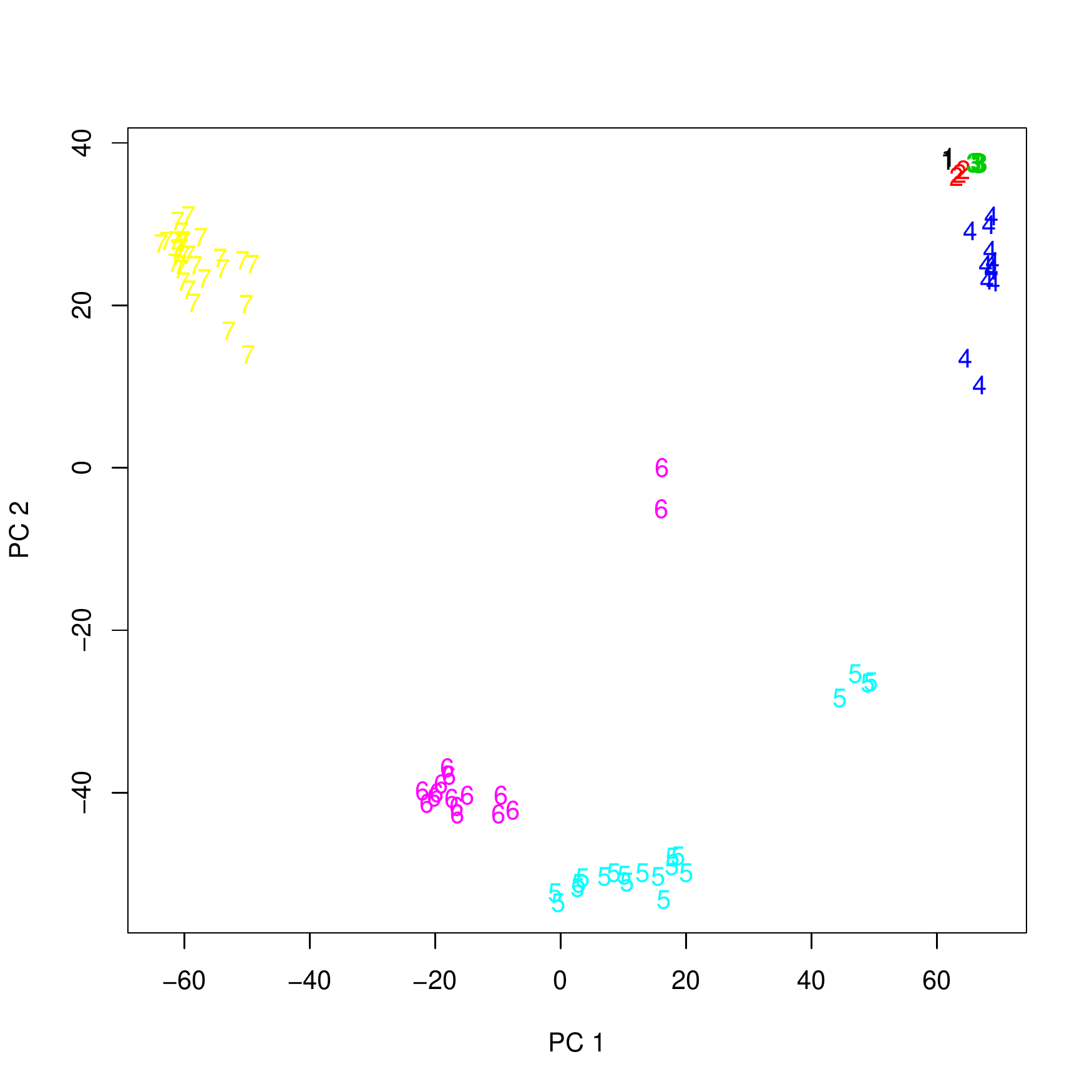}
 \includegraphics[width=0.48\textwidth]{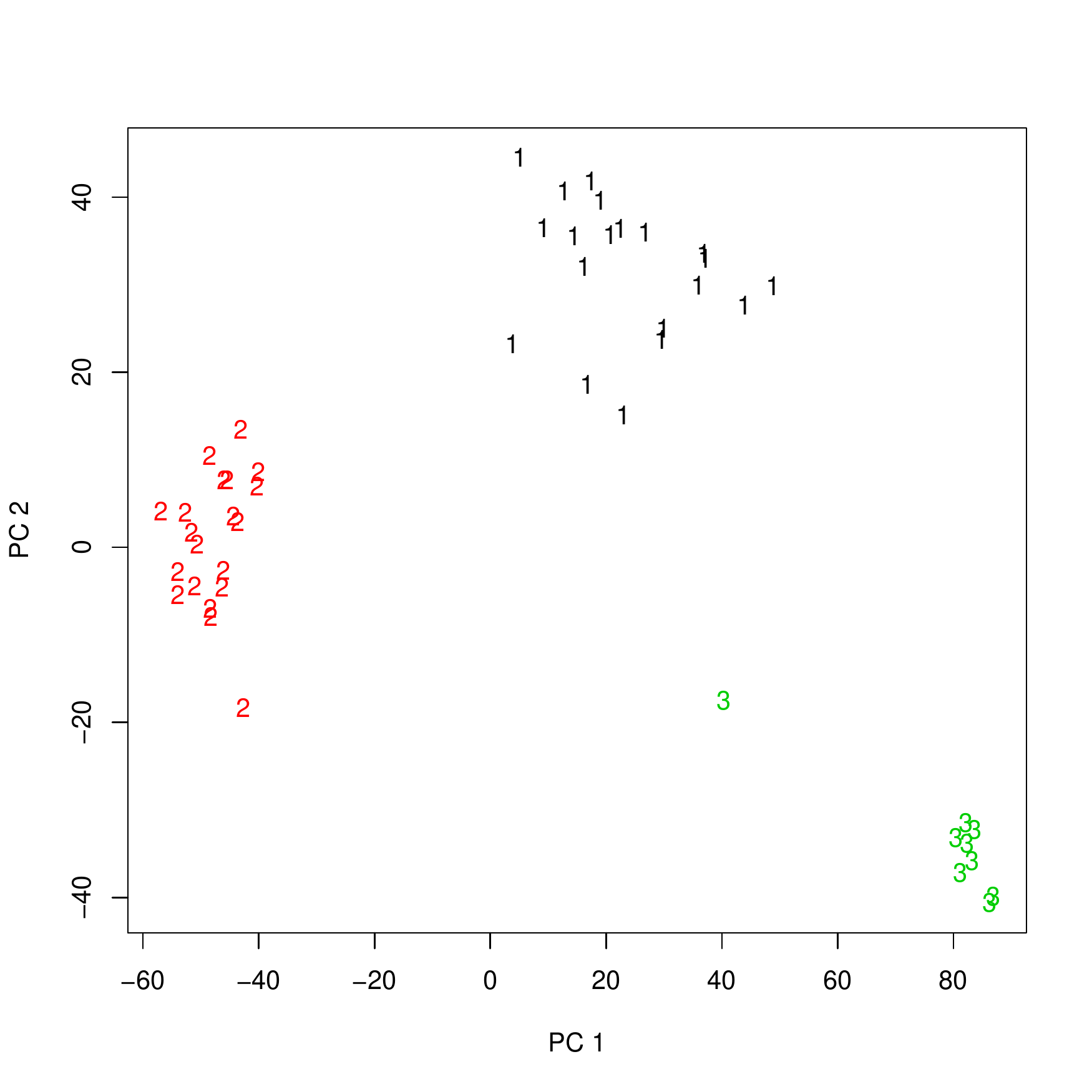}
\\
\includegraphics[width=0.48\textwidth]{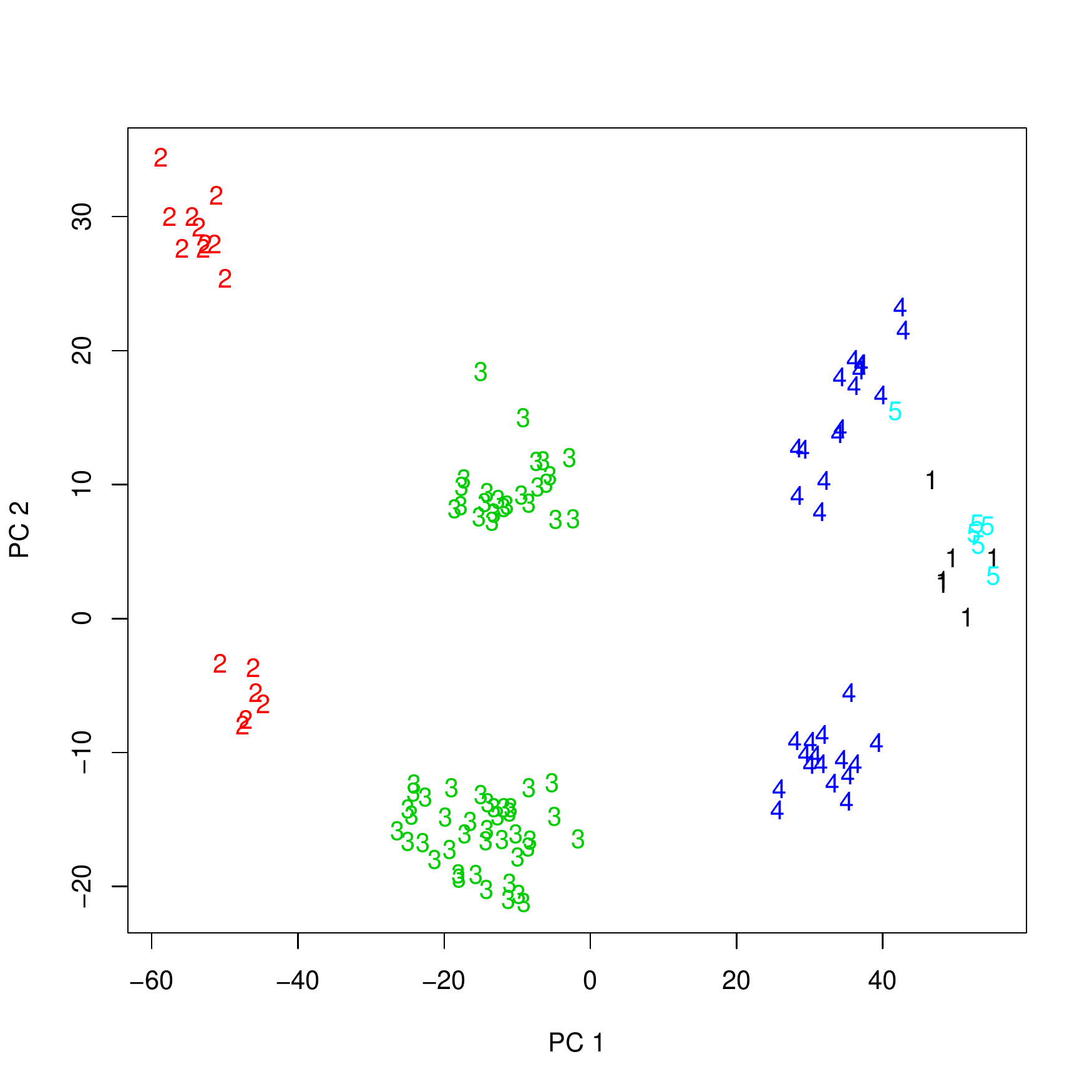}
  \includegraphics[width=0.48\textwidth]{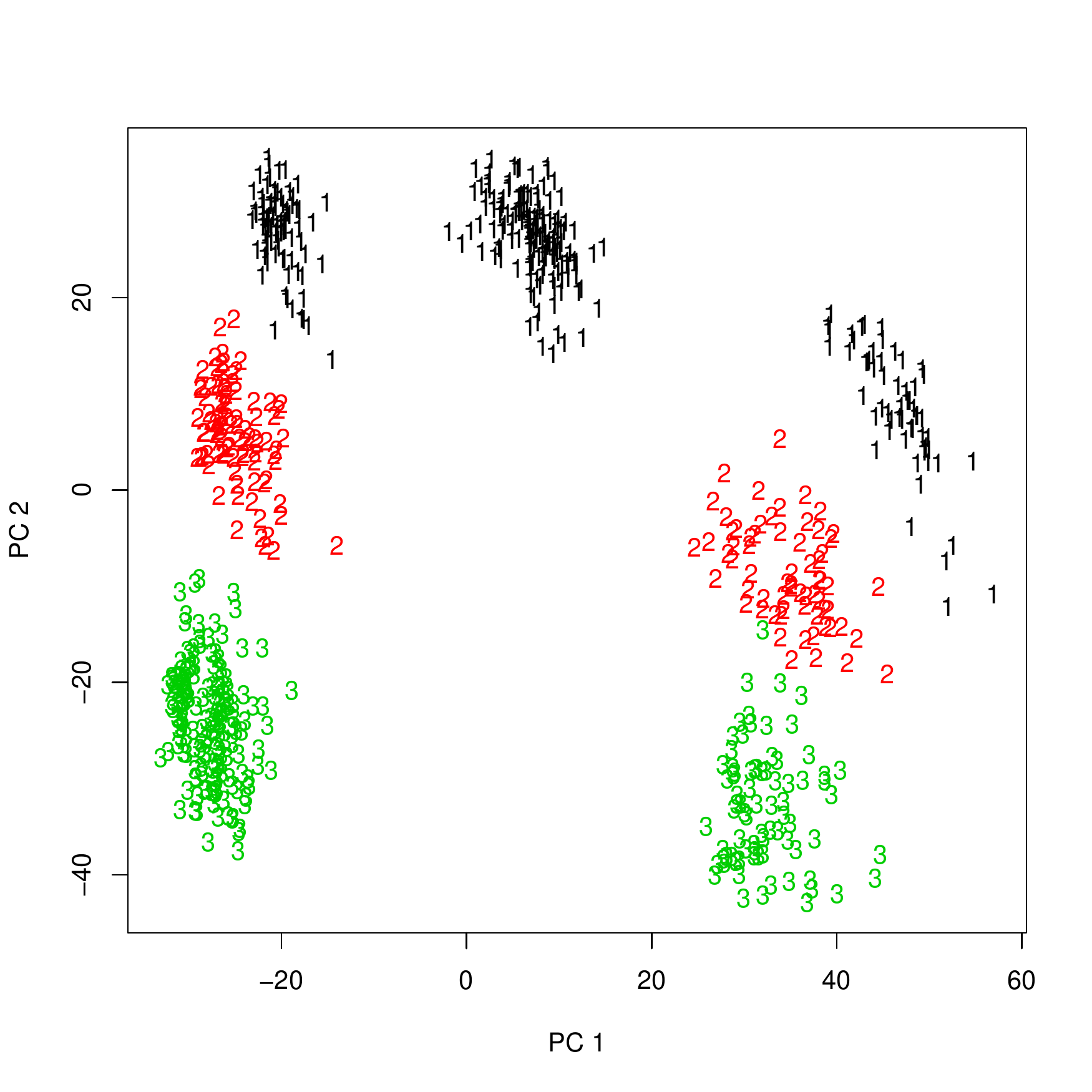}
\caption{Plots of first two PCs with true cell types: (a) \citet{yan2013single} data, (b) \citet{biase2014cell} data, (c) \citet{goolam2016heterogeneity}, (d) \citet{kolodziejczyk2015single} data.}
\label{fscrndata}
\end{figure}

The first two PCs of all data sets are plotted in Figure \ref{fscrndata}. 
It can be seen here is that for all data sets except the one 
of \citet{biase2014cell}, the ``true'' clusters have clearly separated 
subgroups. One could therefore expect a reasonable cluster analysis method to
choose more clusters than  the given ``true'' 
groups, although this does not happen for the data of \citet{yan2013single},
because some ``true'' clusters are not well separated. 
In case of \citet{kolodziejczyk2015single} it is actually known that
there are meaningful sub-populations within each culture condition. 
These are
subsets of the true clusters, so similarity of solutions to the true clusters as
measured by the ARI is still relevant.  
   
\begin{table}[H] \centering 
  \caption{Clustering results for the scRNA-seq data sets of \cite{yan2013single} and \cite{biase2014cell} for all methods included in the comparison with $k$ fixed as the true known $k$, for estimated $k$, and averaged over all $k\in\{2,\ldots,12\}$. Best ARI values in every column are boldfaced.} 
\fontsize{11}{11}\selectfont
\begin{tabular}{l|ccc|cc|c} 
\toprule 
&  \multicolumn{6}{c}{\cite{yan2013single} data}   \\ 
&  \multicolumn{3}{c|}{estimated $k$}   & \multicolumn{2}{c|}{fixed $k=7$} & ave. all $k$    \\ 
Method & ASW  & ARI  & $\widehat{k}$  & ASW & ARI  & ARI    \\
\midrule
k-means & 0.827 & {\bf 0.796}  & 6 & 0.489 & 0.515 & 0.617 \\ 
gap-k-means & 0.803 & 0.791 & 5 & & & \\
PAM &  0.803 & 0.791 & 5 & 0.778 & {\bf 0.894}  & 0.722 \\ 
average &  0.827 & {\bf 0.796} & 6 & 0.766 & 0.795 & 0.671  \\ 
Ward's &  0.827 &  {\bf 0.796} & 6 & 0.778 & {\bf 0.894}  & 0.693 \\ 
BIC-mixture & 0.578 & 0.651 & 8 & 0.766 & 0.795 & 0.663 \\
spectral & 0.827 & {\bf 0.796} & 6 & 0.716 & 0.842 & 0.565 \\ 
PAMSil & 0.827  & {\bf 0.796} & 6 & 0.807 &   0.795 &  0.746\\ 
OSil  & 0.827 & {\bf 0.796}  & 6 &  0.778 & {\bf 0.894} & {\bf 0.750}\\
FOSil  & 0.801 & 0.685  & 3 &  0.592 & 0.632 &  0.644\\
\midrule
&  \multicolumn{6}{c}{\cite{biase2014cell} data}   \\ 
&  \multicolumn{3}{c|}{estimated $k$}   & \multicolumn{2}{c|}{fixed $k=3$} & ave. all $k$    \\ 
Method & ASW  & ARI  & $\widehat{k}$  & ASW & ARI  & ARI   \\
\midrule
k-means & 0.765 & {\bf 1.000} & 3 & 0.765 & {\bf 1.000} & 0.581 \\ 
gap-k-means & 0.765 & {\bf 1.000} & 3 & \\ 
PAM &  0.765 & {\bf 1.000} & 3 & 0.765 & {\bf 1.000}  & 0.575 \\ 
average &  0.765 & {\bf 1.000} & 3 & 0.765 & {\bf 1.000} & 0.683  \\ 
Ward's &  0.765 &  {\bf 1.000} & 3 & 0.765 & {\bf 1.000}  & 0.619 \\ 
BIC-mixture & 0.560 & 0.902 & 4 & 0.765 & {\bf 1.000} & 0.612 \\
spectral & 0.765 & {\bf 1.000} & 3 & 0.765 & {\bf 1.000} & 0.535 \\ 
PAMSil & 0.765  & {\bf 1.000} & 3 & 0.765 &   {\bf 1.000} &  0.682\\ 
OSil  & 0.765 & {\bf 1.000}  & 3 &  0.765 & {\bf 1.000} & {\bf 0.751}\\
FOSil  & 0.765 & {\bf 1.000}  & 3 &  0.765 & {\bf 1.000} &  0.716\\
\bottomrule
\end{tabular} 
\label{tab:yantable}
\end{table}

\begin{table}[H] \centering 
  \caption{Clustering results for the scRNA-seq data sets of \cite{goolam2016heterogeneity} and \cite{kolodziejczyk2015single} for all methods included in the comparison with $k$ fixed as the true known $k$, for estimated $k$, and averaged over all $k\in\{2,\ldots,12\}$. Best ARI values in every column are boldfaced.} 
\fontsize{11}{11}\selectfont
\begin{tabular}{l|ccc|cc|c} 
\toprule 
&  \multicolumn{6}{c}{\cite{goolam2016heterogeneity} data}  \\ 
&  \multicolumn{3}{c|}{estimated $k$}   & \multicolumn{2}{c|}{fixed $k=5$} & ave. all $k$    \\ 
Method & ASW  & ARI  & $\widehat{k}$  & ASW & ARI  & ARI   \\
\midrule
k-means & 0.632 &  0.544 & 5 & 0.632 &  0.544 & 0.493 \\ 
gap-k-means & 0.698 & {\bf 0.571} & 7 & & & \\
PAM &  0.698 &  {\bf 0.571} & 7 & 0.623 &  0.528  & 0.458 \\ 
average &  0.698 &  {\bf 0.571} & 7 & 0.566 &  {\bf 0.842} & {\bf 0.624}  \\ 
Ward's &  0.690 &   0.566 & 7 & 0.624 &  0.543  & 0.518 \\ 
BIC-mixture & 0.598 & 0.432 & 9 & 0.632 &  0.544 & 0.400 \\
spectral & 0.698 &  {\bf 0.571} & 7 & 0.617 &  0.562 & 0.436 \\ 
PAMSil & 0.698  &  {\bf 0.571} & 7 & 0.632 &    0.544 &  0.590\\ 
OSil  & 0.698 &  {\bf 0.571}  & 7 &  0.632 &  0.544 &  0.583\\
FOSil  & 0.661 &  0.522  & 6 &  0.632 &  0.544 &  0.500\\
\midrule
&  \multicolumn{6}{c}{\cite{kolodziejczyk2015single}  data}  \\ 
&  \multicolumn{3}{c|}{estimated $k$}   & \multicolumn{2}{c|}{fixed $k=3$} & ave. all $k$    \\ 
Method & ASW  & ARI  & $\widehat{k}$  & ASW & ARI  & ARI   \\
\midrule
k-means & 0.525 &  0.451 & 7 & 0.389 &  0.295 & 0.336 \\ 
gap-k-means & 0.420 & 0.347 & 12 & & & \\
PAM &  0.528 &  0.524 & 7 & 0.465 &  {\bf 0.434}  & 0.423 \\ 
average &  0.506 &  0.478 & 12 & 0.465 &  0.389 & 0.399  \\ 
Ward's &  0.526 &   {\bf 0.531} & 7 & 0.465 &  0.389  & 0.413 \\ 
BIC-mixture & 0.442 & 0.432 & 9 & 0.415 &  0.147 & 0.418 \\
spectral & 0.499 &  0.359 & 4 & 0.415 &  0.147 & 0.345 \\ 
PAMSil & 0.529  &  {\bf 0.531} & 7 & 0.465 &    0.432 &  {\bf 0.438}\\ 
OSil  & 0.529 &  {\bf 0.531}  & 7 &  0.465 &  0.389 &  0.421\\
FOSil  & 0.528 &  0.525  & 7 &  0.465 &  0.432 &  0.412\\
\bottomrule
\end{tabular} 
\label{tab:gootable}
\end{table}

Tables \ref{tab:yantable} and \ref{tab:gootable} show ASW and ARI results for most of the methods also compared in the simulation study (single linkage yields inappropriate results). We show ASW, ARI, and the estimated number of clusters $\widehat{k}$ for the solutions optimizing the ASW. The ARI here is the most important result, because in reality the assumption will usually be that $k$ is unknown. We also show ASW and ARI values at fixed ``true'' $k$. As this is often not a reasonable ``data analytic'' number of clusters, we also show the average ARI over all $k$ between 2 and 12 in order to investigate whether the methods based on optimizing the ASW for fixed $k$ can find good clusters also at other numbers $k$.

Results differ between data sets. The \cite{biase2014cell} data (Table \ref{tab:yantable}) are easiest to cluster. Almost all methods find the true clustering and estimate $k=3$ correctly, except the BIC-mixture. Some more differentiation occurs at the averaged ARI over all $k$. OSil yields the best result here, followed by FOSil. 

For the \cite{yan2013single} data (Table \ref{tab:yantable}), OSil is best in all respects, but for estimated $k$ many other methods find the same solution. FOSil drops somewhat in quality compared with the other ASW-optimizing methods. 

For the \cite{goolam2016heterogeneity} data (Table \ref{tab:gootable}), most methods estimate a higher $k$ than the true $k=5$ here, as was to be expected. $k$-means with ASW delivers $\widehat{k}=5$, but the corresponding clustering is worse than most others in terms of the ARI. For estimated $k$, the best clustering is found by OSil, PAMSil, PAM, average linkage, $k$-means with the gap statistic, and spectral clustering.
 Regarding fixed $k$, average linkage (which is up with PAMSil and OSil regarding estimated $k$) produces a solution that is far superior to everything else, also lifting its average ARI over all $k$ to the top spot. PAMSil and OSil follow on the next positions.    

For the \cite{kolodziejczyk2015single}  data (Table \ref{tab:gootable}) and estimated $k$, OSil, PAMSil, and Ward do best. The BIC-mixture and gap/$k$-means are much worse. For fixed $k$, PAM is best, only very narrowly over PAMSil and FOSil. There are three clustering solutions with almost the same ASW 0.465 (rounded) at the true $k=3$, namely the one found by PAM, the one found by PAMSil and FOSil, and the third one by OSil, average linkage and Ward. All these are clearly better than the solutions found by $k$-means, model-based clustering, and spectral clustering. Regarding the average ARI over all $k$,  PAMSil is best, followed by PAM and OSil. This is the largest of the data sets, and FOSil, which is meant for larger data sets, loses less quality here compared with OSil than for the smaller data sets, and is substantially faster. 

Considering ASW-values, as in the simulation study, PAMSil is sometimes better and sometimes worse than OSil. Overall, in terms of optimizing the ASW, PAMSil does a good job, and in practice one may run them both and then pick the one that yields the better ASW. The ASW does a generally good job picking the number of clusters; almost all ARI values with $k$ estimated by optimizing the ASW are clearly better than the ARI at true $k$ or averaged over $k$, and the ASW seems to be far more suitable here than model-based clustering with the BIC, probably because the BIC tries to approximate non-Gaussian clusters by more than one Gaussian component. The BIC always yields the largest $\widehat{k}$ here leading to a usually weak ARI.

OSil and PAMSil show the best performances overall. Regarding estimated $k$ they are the only methods that find the best solution for all four data sets, with some good results for average linkage, Ward, and PAM. These three each miss the best solution just once, but do occasionally much worse averaged over all $k$.

\subsection{Species delimitation of Veronica plants} \label{sveronica}
\begin{figure}[H]
\centering
\includegraphics[width=0.48\textwidth]{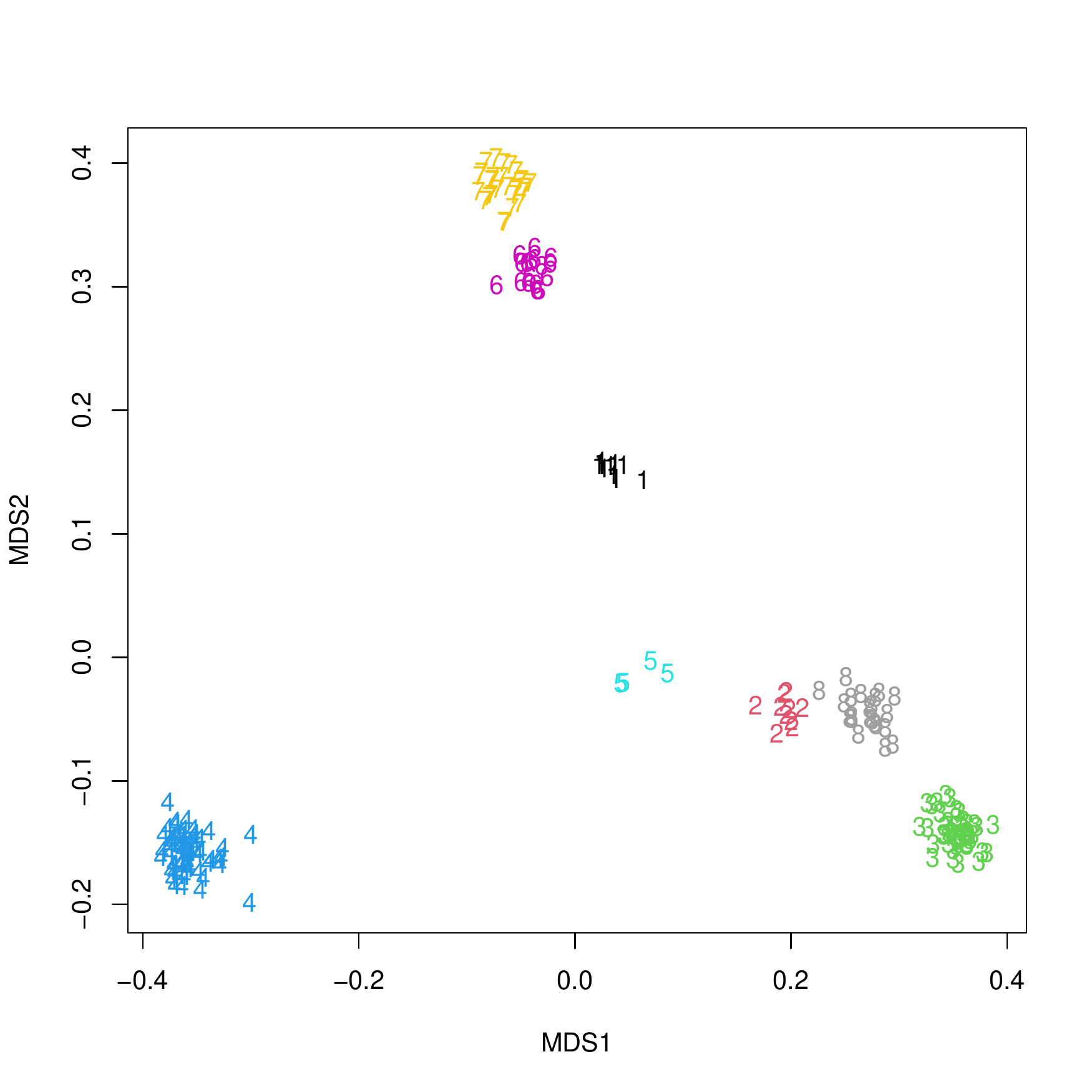}
\includegraphics[width=0.48\textwidth]{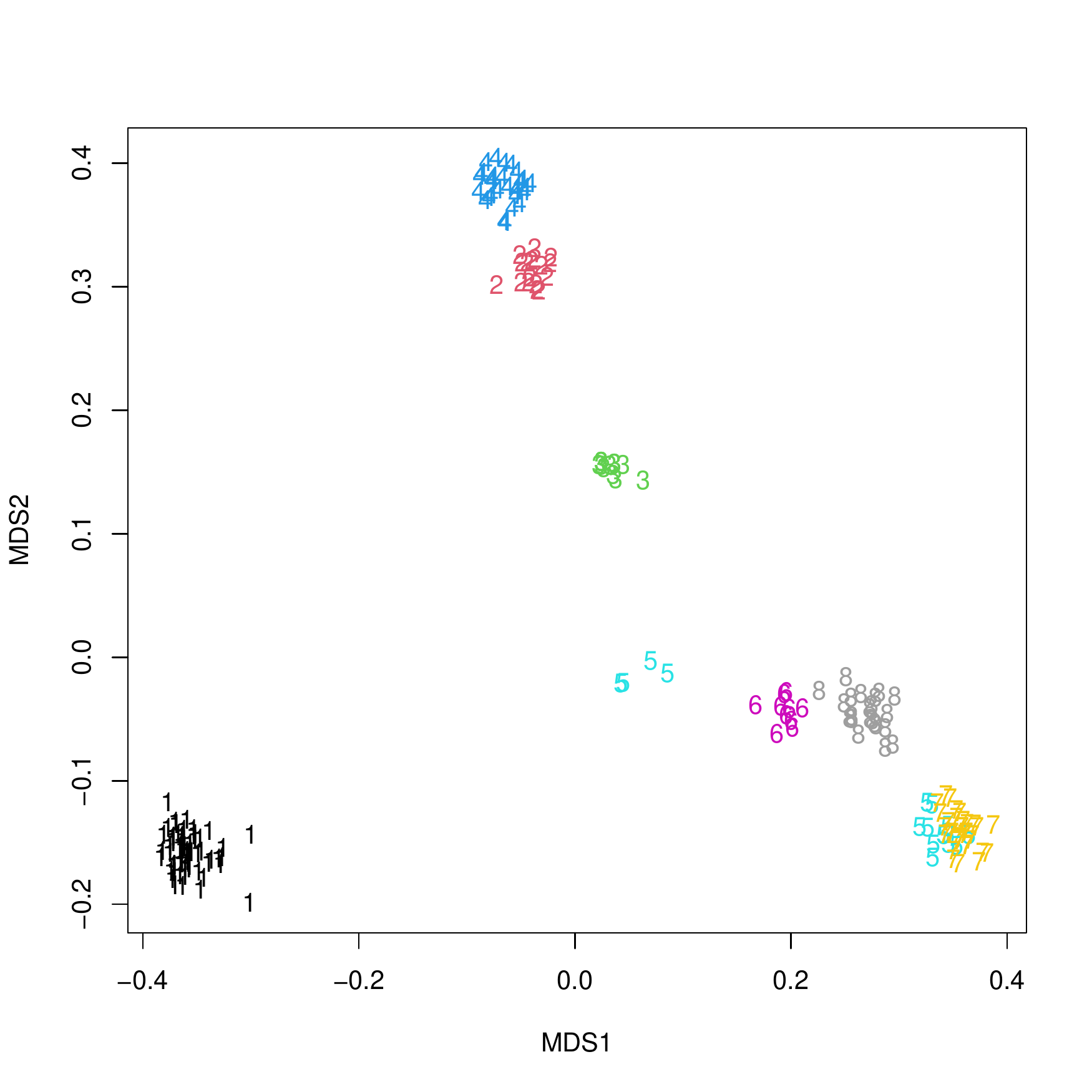}
\caption{2-dimensional multidimensional scaling representation of the Veronica
plants data. Left side: ``True'' species, identical to the solution found by OSil and PAMSil. Right side: Solution found by PAM for 8 clusters.}
\label{fveronica}
\end{figure}

The Veronica data set analysed here is from \citet{MODAERR04}. It gives 
genetic information about 207 individual Veronica plants of sub-genus {\it Pentasepalae} from the Iberian peninsula and Morocco. 
The aim of clustering these is to discover and delimit different species of
such plants. The plants are characterized by 583 variables. These contain
genetic information, which was obtained using AFLP-technology 
(``amplified fragment length polymorphism''). This detects the presence or
absence of certain characteristics (``markers'') of DNA fragments (AFLP
bands from 61 to 454 bp). The 
variables can take 
the values 0 (absence) and 1 (presence). 
As joint presences are the key information and far more relevant than joint 
absences, we computed a Jaccard distance matrix (see, e.g., \citet{Shi93})
on the individuals. 
Involving also further information, \citet{MODAERR04} gave a ``true'' species 
classification into eight species.
Figure \ref{fveronica} shows a 
2-dimensional classical multidimensional scaling (MDS) representation of the 
data. The left side shows the true species, which look very much in line 
with the data. Running OSil and PAMSil on the data, estimating the number of
clusters, delivers $\widehat{k}=8$ clusters that are exactly identical to those 
true clusters. The ASW picks the same solution out of the average linkage and
complete linkage dendrogram. As this is distance data, mclust and 
k-means are not available in their standard form. The ASW was 
in \citet{kaufman1990finding} proposed for use together with PAM, but with PAM 
it suggests 7 clusters, merging somewhat counter-intuitively true species
5 and 8. The reason is that the 8-cluster PAM solution, shown on the right side
of Figure \ref{fveronica}, is even worse in terms of the ASW. 
The PAM objective function for this solution is better than for the 
true 8-cluster solution, so the problem is not that the PAM algorithm would
not find a global optimum. Inspecting individual distances, it turns out that 
according to the PAM objective function it is better to fit the 48 members 
of true species 3 by two centroids, sacrificing the only 4 members of
true species be merging them with a part of true cluster 3.
   
Optimizing the ASW also for fixed number of clusters $k=8$ stops this from 
happening, because the large distance within PAM-8 cluster 5 spoils the 
homogeneity ($a(i)$) values in the silhouette, whereas separating true 
clusters 5 and 3 is good for the separation ($b(i)$).
The OSil solution is not only better in line with the ``true'' species, but also
with the impression given by the MDS plot, and going through individual
distances.
 
This is an example how optimizing the ASW can be beneficial, particularly 
compared to PAM, for a non-Euclidean distance. Other methods (e.g., average and
complete linkage) contain the true 8-species solution in the dendrogram, but
require the ASW to pick $k=8$.

\subsection{France rainfall data clustering} \label{sfrance}

\begin{figure}[H]
\centering
\includegraphics[width=0.48\textwidth]{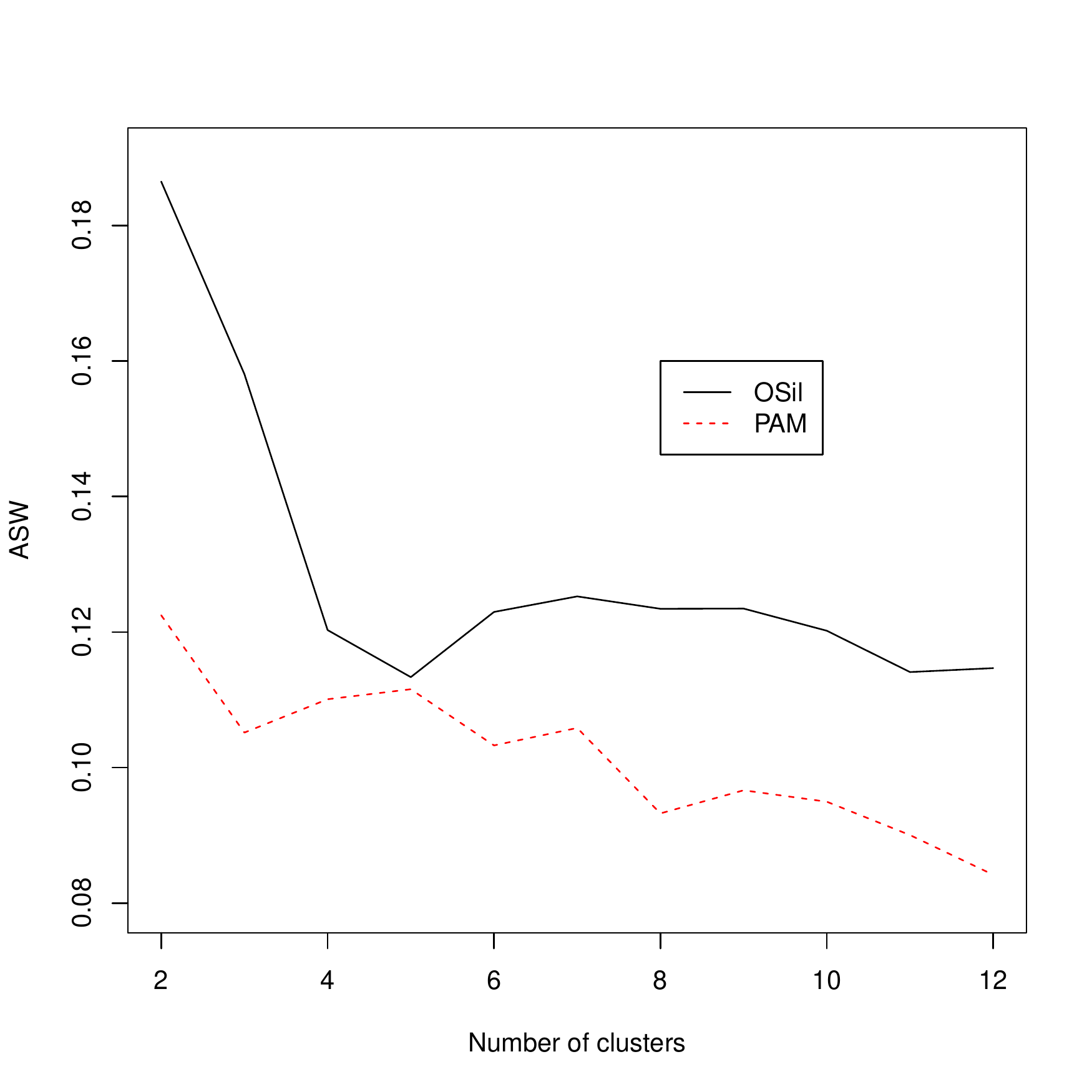}
\caption{France rainfall data: ASW for clusterings from 2 to 12 clusters by OSil
and PAM.}
\label{ffranceasw}
\end{figure}

Finding spatial or temporal patterns in climate data sets based on statistical techniques is of crucial importance for climatologists. 
The data analysed here is taken from \citet{bernard2013clustering}, who clustered 92 French weather stations based on rainfall for the three months of fall, September to November from 1993 to 2011, considering weekly maxima of hourly precipitation, resulting in time series of length 288. Based on subject matter considerations, they proposed a specific
distance measure, the F-madogram, and then ran a PAM clustering on the data. 
They used the ASW in a rather exploratory manner. For PAM, the optimum number of
clusters is $k=2$ (see Figure \ref{ffranceasw}), however the authors are also 
interested in more than two clusters, and they also show and interpret 
solutions for $k=5$ and $k=7$, which are local optima of the 
ASW. As Figure \ref{ffranceasw} shows, OSil yields consistently higher values of
the ASW. The two-cluster solution of OSil just separates two 
outlying weather stations from the 
rest, which is not very useful. 

This points to a weakness of the ASW, which has
a tendency to favor $k=2$ or a low $k$ if one or more clearly separated 
clusters (potentially small) can be found,
ignoring less strongly separated 
structure elsewhere, see Section \ref{soutliers}. 

\begin{figure}[H]
\centering
\includegraphics[width=0.48\textwidth]{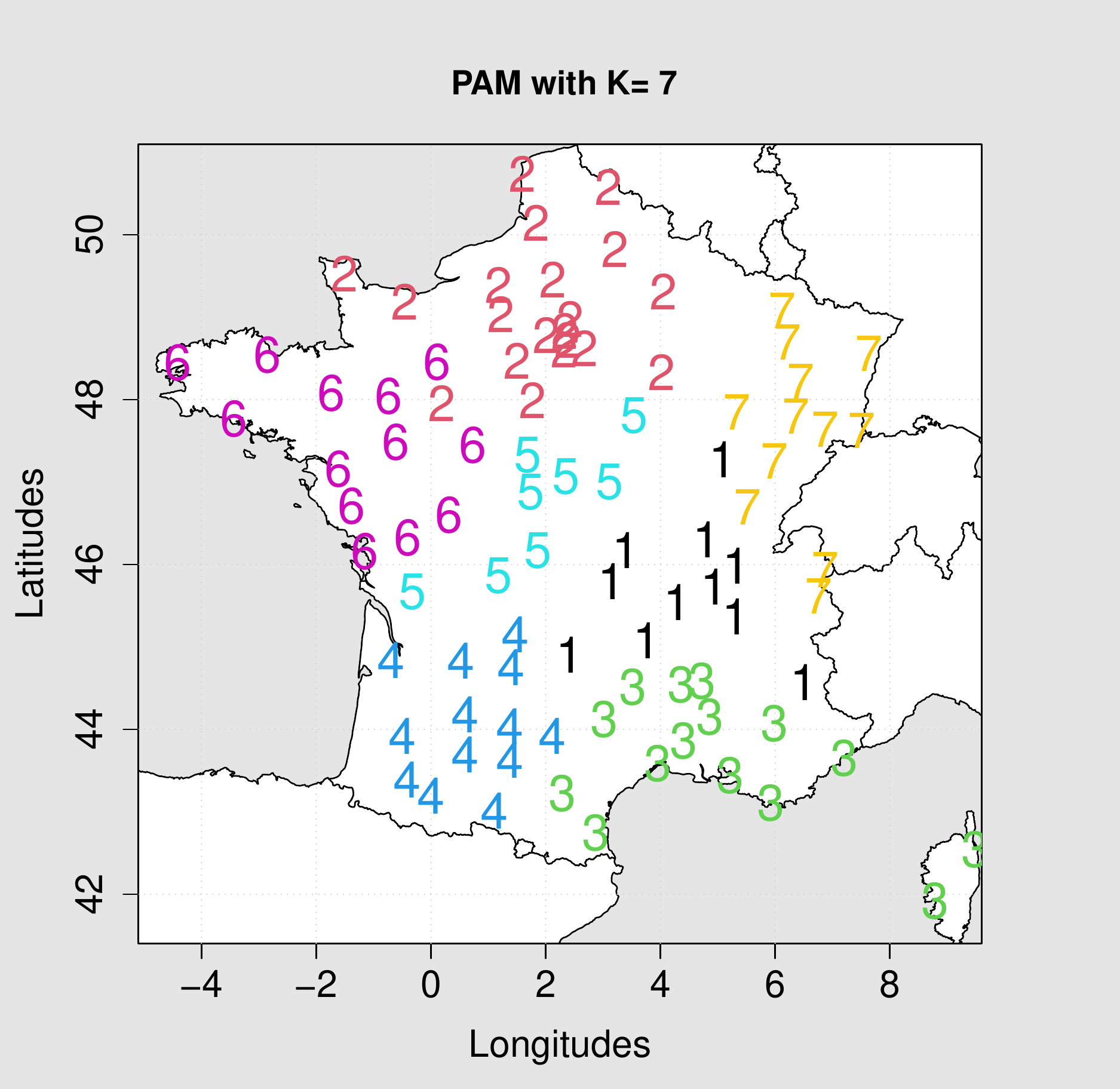}
\includegraphics[width=0.48\textwidth]{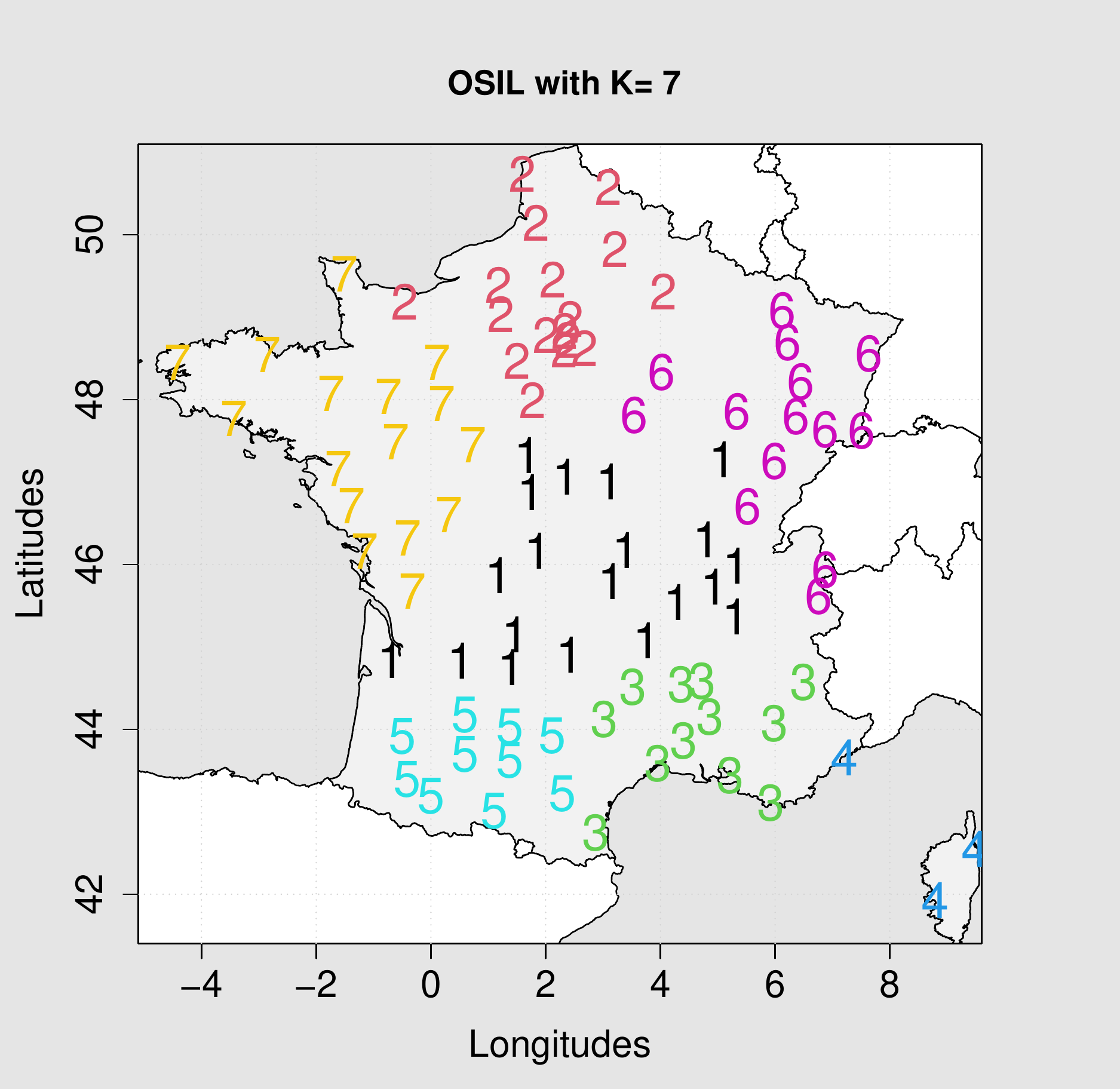}
\includegraphics[width=0.48\textwidth]{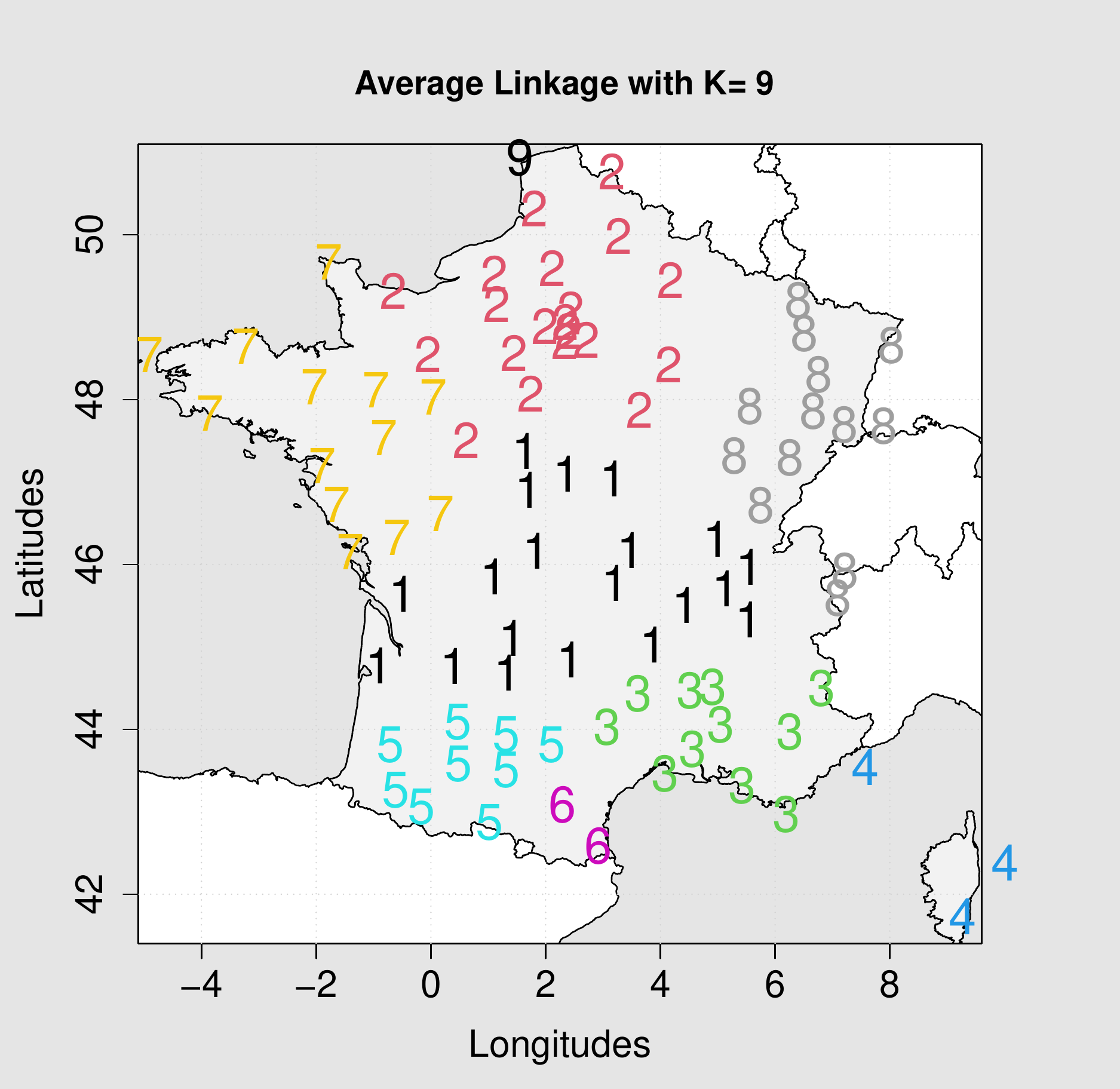}
\caption{France rainfall data: Clusterings for PAM and OSil ($k=7$) and average linkage ($k=9$).}
\label{ffrancecluster}
\end{figure}

A simpler approach is to look for local optima, 
and OSil delivers a local optimum at $k=7$ (see Figure \ref{ffranceasw}). 
Figure \ref{ffrancecluster} shows the 7-cluster solutions for PAM (as used in
\citet{bernard2013clustering}), OSil, and (for comparison) a local ASW optimum
at $k=9$ of average linkage. There is no true clustering given for these data,
but one can look at individual distances to assess the different clusterings.

OSil cluster 4 has just three stations, two of them on Corsica, with average
within-cluster distance (awcd) 0.095. OSil cluster 3 has an awcd of 0.091.
PAM cluster 3 is mainly a merger of these two, and has a much larger awcd
 of 0.101. There is an average distance of 0.115 between OSil clusters 3
and 4, whereas the average distance between PAM clusters 3 and 1 is just 0.111.
This means that OSil clusters 3 and 4 actually seem well separated and leaving
them separated is more convincing that merging them, as PAM does. On the other
hand, OSil cluster 1 is much bigger than the roughly corresponding 
PAM clusters 1 and 5. The awcds of these two PAM clusters are 0.085 and 0.086,
respectively. OSil cluster 1 has awcd 0.093, which is admittedly bigger, 
however three of the seven PAM clusters and only two OSil clusters
have awcds larger than that (the largest awcd of any cluster of both 
clusterings is 0.101 of PAM cluster 3). 
The average distance to the closest
different cluster for OSil cluster 1 is 0.108, and not only is this larger than
the average distance 0.098
between PAM cluster 1 and 5, also both are closer on average (0.107 and 
0.104) to the closest of the remaining five PAM-clusters. Overall the OSil
clustering has a smaller average within-cluster distance, and a larger
average between-cluster distance, and it looks overall more
convincing. OSil focuses more on homogeneity and
separation, whereas PAM focuses more on representation by the best centroid,
which is arguably not that important in this application. 

We ran some other distance-based clustering methods on these data. PAMSil, as
usually, performed similar to OSil, with an ASW optimum at $k=2$ and a local
optimum at $k=7$, however the ASW maximum found by OSil was not found and
PAMSil's solution is a bit different and slightly worse in ASW. All tried out
methods had their ASW optimum at $k=2$. Figure \ref{ffrancecluster} shows
the local ASW optimum  at $k=9$ for average linkage (only local optimum for
$k\le 12$, which was the largest $k$ we tried). The smaller clusters 6 
(2 stations) and 9 (just one station) in
this solution have average distances to the closest other cluster of 0.113 and 
0.108, smaller than the corresponding value between OSil clusters 3 and 4, so
they are hardly strongly outlying, and arguably not that useful. Complete and
single linkage do not yield very convincing solutions either.

Lacking ``true'' cluster information (as is typically the case 
in real applications), such arguments elaborate 
in what sense the OSil method can achieve something valuable for these data that
is not achieved by the other methods. 
This is only achieved after 
accounting for a weakness of the ASW for estimating $k$ (for which it is
regularly used), namely that it can 
get stuck at $k=2$ because of large within-cluster distances that hide smaller
but still relevant separation at a higher level of $k$. In real data analysis,
higher local optima should also be explored, be it with OSil,
or be it where the ASW is used with other methods for choosing $k$.

\section{Conclusion}\label{sconclusion}
We introduce the OSil and FOSil methods for optimum ASW clustering. 
The ASW is shown to fulfil the desirable axioms for clustering quality measures proposed by \citet{ben2009measures}. From our experiments, the ability of OSil to find ``true'' clusters is good, but there are exceptions. It was the strongest method for DGPs containing clearly separated clusters with differing spreads and sizes. For several further models, including those with higher dimensionality, it performed well, and in line with most other methods. Results for the scRNA-seq data were overall best.
An issue, highlighted in Sections {\ref{soutliers} and \ref{sfrance}, 
but also present in some simulations, is that the ASW as an estimator of the number of clusters $k$ can be tempted to choose a too low $k$ if this allows for very large distances between ``neighboring'' clusters, which can hide structure that is still meaningful but characterized by somewhat lower between-cluster distances. This is a problem not only with OSil, but with the widespread general use of the ASW as a criterion to estimate $k$. It is advised to consider locally optimal values of $k$ regarding the ASW on top of the global optimum where the global optimum is at $k=2$ or very low. A bootstrap 
scheme to estimate the number of clusters with the ASW correcting for bias in 
favor of low $k$ has been proposed in
\citet{hennig15paramboot}.
On the other hand, the applications to non-Euclidean data show that OSil can achieve sensible results where PAM and other methods have difficulties. Also the ASW did better finding good clusterings at estimated $k$ for the scRANA-seq data than the BIC with a Gaussian mixture model.
  
The PAMSil algorithm by \citet{van2003new} is a good approximation to OSil, often finding the same optimum ASW, sometimes worse, sometimes even better. For larger data sets, the FOSil algorithm based on subsetting has been proposed, which however occasionally results in considerable quality loss. As long as the optimisation of the ASW is of interest in its own right, FOSil still usually achieves a better ASW-value than other clustering methods apart from the slower OSil and PAMSil. 

\section*{Software} 
An R package is available at the first author's Github site at \url{https://github.com/bfatimah}. Code for PAMSil was built on the standalone C-code by \citet{van2003new}.

\section*{Acknowledgement}
The research of the first author was funded by the Commonwealth Scholarship Commission. 


\section*{Appendix: Definition of DGPs} \label{dgpdefs}
Let $N_p(\mu_p, \Sigma_{p\times p})$ denote the $p$-variate Gaussian distribution with mean $\mu_p$ and covariance matrix $\Sigma_{p\times p}$. Let $SN(\zeta, \omega, \alpha, \tau)$ denote a skew Gaussian univariate distribution with $\zeta, \omega, \alpha, \tau$ as location, scale, shape, and hidden mean parameters, respectively. Let $\mathds{U}(a, b)$ denote the uniform distribution defined over the continuous interval $a$ and $b$. Let $t_v$ denote Student's $t$ distribution with $v$ degrees of freedom. Let $t_r(\nu)$ denote the non-central $t$ distribution with $r$ degrees of freedom and non-centrality parameter $\nu$. $\textrm{Gam}(\alpha, \beta)$ denotes the Gamma distribution, where $\alpha$ and $\beta$ are shape and rate parameters, respectively. $\textrm{NBeta}(v_1, v_2, \lambda)$ denotes the non-central Beta distribution of Type I with $v_1, v_2$ the two shape parameters, and non-centrality parameter  $\lambda$. Let  $\textrm{Exp}(\lambda)$ denote the Exponential distribution with rate parameter $\lambda$.  $\mathds{F}_{(v_1, v_2)}(\lambda)$ denotes the non-central $F$ distribution with degrees of freedom $v_1, v_2$ and non-centrality parameter $\lambda$.  $\mathds{W}(\tau, \zeta)$ denotes the Weibull distribution with $\tau, \zeta$ as shape and scale parameter, respectively. $I_p$ denotes identity matrix of order $p$.\\
\noindent{\bf{DGP 1: }} 2 clusters in 2 dimensions: The clusters were generated from  two independent Gaussian random variables. Cluster  1 has mean (0, 5) and covariance matrix $0.7^2I_2$, cluster 2 has mean (0, 5) and covariance matrix  $0.1^2I_2$.\\
\noindent{\bf{DGP 2: }}  3 clusters in 2 dimensions: The observations in each of the three clusters were generated from independent Gaussian random variables with mean  (0, 0) and covariance matrix $0.7^2I_2$  for cluster 1, mean
(-2, 0) and covariance matrix $0.1^2I_2$  for cluster 2, and  mean (2,  0) and covariance matrix $0.1^2I_2$  for cluster 3. \\
\noindent{\bf{DGP 3: } } 4 clusters in 2 dimensions: Cluster 1 was generated from two independently distributed non-central variables $t_7(10)$ and $t_7(30)$. Cluster 2 was generated from  independent Gaussian distribution having mean  (2, 2) with covariance matrix $\Sigma = \begin{bmatrix}
 4 &  0\\
  0 & 16 
\end{bmatrix}$.
Cluster 3 was generated from $\mathds{U}(10, 15)$ along both dimensions independently.  Cluster 4 was generated from independent Gaussian distributions  with means  (20, 80) and  covariance matrix   $\Sigma = \begin{bmatrix}
 1 &  0\\
  0 & 4 
\end{bmatrix}$. \\  
\noindent{\bf{DGP 4: }}  5 clusters in 2 dimensions: Variables are independent within clusters. The clusters are generated  from $\chi^2_7(35)$ and $\chi^2_{10}(60)$ (cluster 1); $\mathds{F}_{(2, 6)}(4)$ and  $\mathds{F}_{(5, 5)}(4)$ (cluster 2);   $N(100, 16)$ and $N(0, 16)$ (cluster 3);  $t_{40}(100)$ and $t_{35}(150)$ (cluster 4); $SN(20, 2, 2, 4)$ and $SN(200, 2, 3, 6)$ (cluster 5).\\
\noindent {\bf{DGP 5: }} 6 clusters in 2 dimensions: Independent variables within cluster;  $\textrm{Exp}(10)$ in both dimensions (cluster 1); $\textrm{NBeta}(2, 3, 220)$ and $NBeta(2, 3, 120)$ (cluster 2);  $\mathds{W}(10, 4)$ across both dimensions (cluster 3);  $\textrm{Gam}(15, 2)$ and $\textrm{Gam}(15, 0)$ (cluster 4); $\mathds{U}(-6, -2)$ in both dimensions (cluster 5);  $SN(5, 0.6, 4, 5)$ and $SN(0, 0.6, 4, 5)$ (cluster 6).\\
\noindent {\bf{DGP 6: }} \\
5 clusters in 5 dimensions are generated from multi-variate Gaussian distributions. 
 Cluster 1 is centred at (0, 0, 0, 0, 0) with  
 $\Sigma=
\begin{bmatrix}
  9 &  &   &   &   \\
  1  & 17 & &   &  \\
 1  & -1.4 & 12 & &  \\
  0.4 &  0.6 & 0.5 & 2 & \\
-1.2 & -1.6 & -1.4 & -0.6 & 16\\
\end{bmatrix}$.\\
Cluster 2 is centred at (5,  10,  3,  7, 6) with 
$\Sigma=
\begin{bmatrix} 
  1  & \\
  0.3 & 1 &   \\
  0.3 & -0.3 & 1   &  \\
  -0.3&   0.3 & 0.3 & 1  & \\
 -0.3& -0.3 & -0.3 & -0.3 & 1  \\
\end{bmatrix}$.\\
Cluster 3 is centred at (15,  70, 50, 55, 80) with 
$\Sigma=
\begin{bmatrix} 
  25 & \\
   3  & 9 &   \\
   4  & -2.4 & 16  &  \\
  -1  &  -0.6 & 0.8 & 1  & \\
 -7  & -4.2 & -5.6 & -1.4 & 49 \\
\end{bmatrix}$.\\

Cluster 4 is centred at (70, 80, 70, 70, 70) with 
$\Sigma=
\begin{bmatrix}
  5 & \\
  0.21 & 0.9 &   \\
  0.28 & -0.24 & 1.6 &  \\
 -1.57 &  0.19 & 0.25 &1  & \\
 -1 & -1.89 & -0.56 & -0.44 &4.9 \\
\end{bmatrix}$.\\

Cluster 5 is centred at (55, 55, 55, 55, 55) with 
$\Sigma=
\begin{bmatrix}
  2 & \\
  0.85 & 9 &   \\
  0.49 & -0.52 & 3  &  \\
  -0.42 & 0.6  & 0.17 & 1 & \\
 -0.28 & -0.6 & -0.69 & -1.8 & 4\\
\end{bmatrix}$.\\

\noindent {\bf{DGP 7: }}\\
10 cluster in 500 dimensions. Data for each cluster were drawn from normal distributions with cluster means -16, -13, -10, -6, -3, 3, 6, 10, 13, 21. Within-cluster variances were randomly drawn from $0.005^2, 0.1^2, 0.2^2, 0.3^2, 0.4^2$. The data were copied to all 500 dimensions.\\
\noindent {\bf{DGP 8: }}\\
 7 clusters in 60 dimensions with 500 observations:  
This is a data structure designed by \citet{van2003new} to simulate gene expression profiles like structure for three distinct types of cancer patients' populations. Suppose that in reality there are 3 distinct groups 20 patients each corresponding to a cancer type.  Three multivariate normal distributions were used to generate 20 samples each having different mean vectors.  For the first multivariate distribution (first cancer type), the first 25 dimensions(genes) are centred at $\log_{10}(3)$, dimensions 26-50 are centred at $(-\log_{10}(3))$,  the remaining 450 dimensions are centred at 0.  For the second multivariate distribution (second cancer type), the first 50 dimensions(genes) are centred at $0$, the next 25 dimensions (51-75) are centred at $\log_{10}(3)$,  dimensions 76-100 are centred at $(-\log_{10}(3))$,  and  the remaining 400 dimensions are also centred at 0. For the third multivariate distribution (third cancer type) the first 100 dimensions(genes) are centred at $0$, dimensions 101-125 are centred at $\log_{10}(3)$,  dimensions 126-150 are centred at $(-\log_{10}(3))$  and dimensions 151-500 are also centred at 0.  The three multivariate distributions have diagonal covariance matrix with diagonal elements as $(\log(1.6))^2$. Note that the described data has 20 samples each of 3 types of cancer patients each containing 500 genes. The purpose here is to cluster genes, not patients. Therefore, the transpose of the data is required to transfer it to the standard format and the number of clusters to seek are 7 in 60 dimensions of 500 observations (\cite{van2003new} seem to explain only 400 genes, but this looks like a typo).\\  
\noindent {\bf{DGP 9: }}\\
3 clusters in 1000 dimensions. Each cluster contains 40 realisations from standard Gaussian distributions with each of first 100 coordinates centred at -3, 0, and 3 respectively. The remaining coordinates of all clusters have mean 0. All the clusters have $I_{1000}$ covariance matrices.

\end{document}